\documentclass[lettersize,journal]{IEEEtran}
\IEEEoverridecommandlockouts
\usepackage{cite}
\usepackage{algorithmic}
\usepackage{graphicx}
\usepackage{textcomp}
\usepackage{array}
\usepackage{stfloats}
\usepackage{url}
\usepackage{verbatim}
\usepackage{times}
\usepackage{epsfig}
\usepackage[compact]{titlesec}
\usepackage{amsmath}
\usepackage{amssymb}
\usepackage{amsfonts}
\usepackage{booktabs}
\usepackage{fixltx2e}
\usepackage{colortbl}
\usepackage[vlined,ruled]{algorithm2e}
\usepackage{xcolor}
\usepackage{color}
\usepackage{soul}
\usepackage{multirow}
\usepackage{float}
\usepackage{balance}
\usepackage{makecell}
\usepackage{caption}
\usepackage{subcaption}
\usepackage{lipsum}
\usepackage{rotating}
\usepackage{adjustbox}
\usepackage{pdflscape}
\usepackage{siunitx}
\usepackage{lscape}
\usepackage{pifont}
\usepackage{blindtext}
\usepackage{tabu}
\usepackage[T1]{fontenc}
\usepackage[utf8]{inputenc}
\usepackage{flexisym}
\usepackage{forest}
\usepackage{longtable}
\usepackage[multiple]{footmisc}
\usepackage{ctable}
\usepackage{epstopdf}
\usepackage{fixfoot}
\usepackage{multicol}
\usepackage{tabularray}
\usepackage{tabularx}
\usepackage[most]{tcolorbox}
\usepackage{xspace}
\usepackage{academicons}
\usepackage{soul, todonotes}

\makeatletter
\newcommand{\linebreakand}{%
\end{@IEEEauthorhalign}
\hfill\mbox{}\par
\mbox{}\hfill\begin{@IEEEauthorhalign}
}
\makeatother

\newcommand{\xmark}{\ding{55}}%

\newcommand\mycomment[1]\null

\makeatletter
\DeclareRobustCommand\onedot{\futurelet\@let@token\@onedot}
\def\@onedot{\ifx\@let@token.\else.\null\fi\xspace}
\def\eg{\emph{e.g}\onedot} 
\def\ie{\emph{i.e}\onedot}

\def\etal{\emph{et al}\onedot}
\makeatother

\def\BibTeX{{\rm B\kern-.05em{\sc i\kern-.025em b}\kern-.08em T\kern-.1667em\lower.7ex\hbox{E}\kern-.125emX}}

\begin{document}

\IEEEoverridecommandlockouts\IEEEpubid{\makebox[\columnwidth]{
\begin{tabular}[t]{l}
\\\\\\
\copyright This work has been submitted to the IEEE for possible publication.\\
Copyright may be transferred without notice, after which this version may no longer be accessible.\\
Hence, it will be replaced by the accepted version.
\end{tabular}
} \hspace{\columnsep}\makebox[\columnwidth]{ }}

\title{GTA-HDR: A Large-Scale Synthetic Dataset\\ for HDR Image Reconstruction}

\author{
Hrishav Bakul Barua$^{\star}$,
Kalin Stefanov, \IEEEmembership{Member, IEEE},
KokSheik Wong$^{\dagger}$, \IEEEmembership{Senior Member, IEEE},
Abhinav Dhall, \IEEEmembership{Member, IEEE} and
Ganesh Krishnasamy, \IEEEmembership{Member, IEEE}

\thanks{H. B. Barua is with School of Information Technology, Monash University, and Robotics and Autonomous Systems Group, TCS Research, India (e-mail: hrishav.barua@monash.edu).}
\thanks{K. Stefanov is with Faculty of Information Technology, Monash University, Australia (e-mail: kalin.stefanov@monash.edu).}

\thanks{K. Wong is with School of Information Technology, Monash University, Malaysia (e-mail: wong.koksheik@monash.edu).}
\thanks{A. Dhall is with Flinders University, Adelaide, Australia, and Centre for Applied Research in Data Sciences, Indian Institute of Technology Ropar, India (e-mail: abhinav@iitrpr.ac.in).}
\thanks{G. Krishnasamy is with School of Information Technology, Monash University, Malaysia (e-mail: ganesh.krishnasamy@monash.edu).}
\thanks{$^{\star}$This research is supported by the Global Excellence and Mobility Scholarship, Monash University.}
\thanks{$^{\dagger}$This research is supported, in part, by the E-Science fund under the project: \emph{Innovative High Dynamic Range Imaging - From Information Hiding to Its Applications} (Grant No. 01-02-10-SF0327).}
}

\markboth{IEEE Transactions on Multimedia}{Barua \etal - GTA-HDR: A Large-Scale Synthetic Dataset for HDR Image Reconstruction}

\maketitle

\begin{abstract}
High Dynamic Range (HDR) content (\ie., images and videos) has a broad range of applications.
However, capturing HDR content from real-world scenes is expensive and time-consuming.
Therefore, the challenging task of reconstructing visually accurate HDR images from their Low Dynamic Range (LDR) counterparts is gaining attention in the vision research community.
A major challenge in this research problem is the lack of datasets, which capture diverse scene conditions (\eg, lighting, shadows, weather, locations, landscapes, objects, humans, buildings) and various image features (\eg, color, contrast, saturation, hue, luminance, brightness, radiance).
To address this gap, in this paper, we introduce GTA-HDR, a large-scale synthetic dataset of photo-realistic HDR images sampled from the GTA-V video game.
%GTA-V uses ray tracing technology and offers inbuilt HDR support, providing the capability of photo-realistic rendering of diverse sets of scenes, objects, humans, and landscapes.
We perform thorough evaluation of the proposed dataset, which demonstrates significant qualitative and quantitative improvements of the state-of-the-art HDR image reconstruction methods.
Furthermore, we demonstrate the effectiveness of the proposed dataset and its impact on additional computer vision tasks including 3D human pose estimation, human body part segmentation, and holistic scene segmentation.
%The proposed dataset represents an important contribution that will enable the development of advanced techniques for visually accurate HDR image reconstruction.
The dataset, data collection pipeline, and evaluation code are available at: \texttt{\url{https://github.com/HrishavBakulBarua/GTA-HDR}}.
\end{abstract}

\begin{IEEEkeywords}
High dynamic range imaging, tone mapping, inverse tone mapping, deep learning, machine learning
\end{IEEEkeywords}

\section{Introduction}
\label{sec:introduction}
\IEEEPARstart{H}{igh} Dynamic Range (HDR)~\cite{artusi2019overview}\mycomment{~\cite{fairchild2007hdr,artusi2019overview}} content (\ie., images and videos) has been adopted widely in various domains including  entertainment~\cite{he2022sdrtv}\mycomment{~\cite{chen2021new,he2022sdrtv}}, gaming and augmented/virtual reality~\cite{satilmis2023deep}\mycomment{~\cite{10.1145/3532721.3535566,satilmis2023deep, huang2022hdr,banterle2017advanced,wang2019co}}, medical imaging~\cite{huang2022hdr}\mycomment{~\cite{10.1145/3532721.3535566,huang2022hdr}}, computational photography~\cite{nguyen2023psenet}\mycomment{~\cite{nguyen2023psenet,zhang2017learning,ren2023robust,yang2023lightingnet}}, and robotics/robot vision~\cite{wu2020hdr}.
However, capturing HDR content from real-world scenes is costly and time-consuming.
Therefore, HDR image reconstruction from Low Dynamic Range (LDR) counterparts has been an active area of research in the last several years~\cite{wang2021deep,tiwari2015review,tursun2015state,johnson2015high,wang2022kunet,kinoshita2019scene}.
The literature proposes a multitude of methods for HDR image reconstruction that are gradually shifting from traditional (non-learning) techniques~\cite{luzardo2018fully,kovaleski2014high,huo2014physiological,masia2017dynamic} towards data-driven learning-based, such as Generative Adversarial Networks~\cite{guo2023single}\mycomment{~\cite{raipurkar2021hdr,niu2021hdr,guo2023single}} and Diffusion Models~\cite{dalal2023single}.
%Fig.~\ref{fig:supervised_learning_pipelines} depicts the standard supervised learning pipelines for HDR image reconstruction from LDR images and no-reference quality assessment of reconstructed HDRs.

%\begin{figure}[t]
%\captionsetup[subfigure]{justification=centering}
%\centering
%\subfloat[Reconstruction]{\includegraphics[scale=0.75]{images/update/reconstruction_pipeline}}
%\hfill
%\subfloat[Quality assessment]{\includegraphics[scale=0.75]{images/update/quality_assessment_pipeline}}
%\caption{\textbf{Standard supervised learning pipelines for a) HDR image reconstruction and b) No-reference quality assessment.} \emph{GT: Ground truth, Rec: Reconstructed, Dist: Distorted.}}
%\label{fig:supervised_learning_pipelines}
%\end{figure}

Given that the performance of any data-driven learning-based method for HDR image reconstruction largely depends on the size and diversity of the datasets used for development, there is a significant gap in the publicly available datasets required to advance this research direction.
Specifically, the existing datasets are either: 1) Not sufficiently large~\cite{kalantari2017deep,endo2017deep,eilertsen2017hdr,nemoto2015visual,lee2018deep,prabhakar2019fast,jang2020dynamic}; 2) Not having satisfactory resolution~\cite{zhang2017learning,hdrsky}; 3) Having limited scene diversity~\cite{kalantari2017deep,prabhakar2019fast, jang2020dynamic}; 4) Having limited image variations~\cite{liu2020single,cai2018learning,kim2019deep,jang2020dynamic}; or 5) Absence of ground truth HDR images~\cite{sen2012robust,tursun2016objective,dang2015raise}.
Furthermore, currently, there are no available datasets that adequately address the problem of no-reference HDR image quality assessment, which demands vast collections of ground truth HDR and distorted HDR pairs~\cite{banterle2020nor,banterle2023nor,artusi2019efficient}.
In summary, there is a substantial research gap pertaining to benchmark datasets needed to advance the research on HDR image reconstruction\mycomment{\cite{wang2021deep}}, hence motivating the creation of an appropriate large-scale dataset.
%The thorough analysis of the existing datasets required to advance this research direction reveals the current limitations of the existing datasets justify rendering the generation of an appropriate large-scale dataset vital.

Video games have been used for creation and annotation of various large-scale datasets in diverse computer vision tasks~\cite{zhang2021simulation}\mycomment{~\cite{zhang2021simulation,johnson2016driving}} including 3D human pose and motion reconstruction~\cite{yang2023synbody}\mycomment{~\cite{cai2021playing,yang2023synbody}}, semantic segmentation~\cite{angus2018unlimited}\mycomment{~\cite{richter2016playing,richter2017playing,krahenbuhl2018free,angus2018unlimited}}, 3D scene layout and visual odometry~\cite{richter2017playing}, pedestrian detection and tracking~\cite{fabbri2021motsynth}, object detection and 3D mesh recovery~\cite{hu2021sail}\mycomment{~\cite{hu2021sail,gaidon2016virtual}}\mycomment{, optical flow estimation~\cite{krahenbuhl2018free}\mycomment{~\cite{richter2017playing,krahenbuhl2018free}}}, optical flow and depth estimation~\cite{krahenbuhl2018free}\mycomment{, and 3D mesh recovery~\cite{hu2021sail}}.
Drawing inspiration from the success of various data-driven learning-based methods developed with video game data\mycomment{~\cite{zhang2021simulation,johnson2016driving,cai2021playing,yang2023synbody,richter2016playing,richter2017playing,krahenbuhl2018free,angus2018unlimited,hu2021sail}}, in this paper, we propose GTA-HDR, a large-scale synthetic dataset for HDR image reconstruction, sampled from the photo-realistic (\ie, HDR-10\footnote{https://en.wikipedia.org/wiki/HDR10}\mycomment{\footnote{https://www.adriancourreges.com/blog/2015/11/02/gta-v-graphics-study/}} enabled) game Grand Theft Auto V\footnote{https://www.rockstargames.com/gta-v} (GTA-V) by Rockstar Games\footnote{https://www.rockstargames.com}. 
Many game-inspired datasets for vision applications are constructed utilizing GTA-V.
Previous work has also used other video games including Hitman~\cite{richter2016playing}, Witcher 3~\cite{witcher}, and Far Cry Primal~\cite{xu2017end}\mycomment{~\cite{xu2017end, krahenbuhl2018free}} which contain highly realistic and detailed worlds with high fidelity.
However, those games lack the diversity of scenes, which is the main trait of GTA-V.
%The main characteristics of GTA-V lies in its  diversity scale, photo-realistic appearance, and natural behavior.

%The data is collected using completing the game-playing sequence (at least once) from beginning to end.
%The full flow of the data collection framework is depicted in figure \ref{fig:dataset_collection_pipeline}.
%The GTA-V has in-game HDR-10 settings which we use to curate HDR images from the game sequence.
%No only that, it will also require traveling from one place to another to diversify the dataset.
%We are able to make use of the in-game rich data to collect scenes from different locations consisting of greenery, forest, mountain, sea, beach, city, indoor, \etc.
%We are able to generalize the data by including indoor, urban, rural as well and in-the-wild scenes.
%We also take special care to include different lighting conditions both indoor and outdoor by collecting scenes from different times of day (early morning, day, evening, and late night).
%The weather and season settings of the game give us realistic-looking seasons (summer and winter) and conditions such as snowy, rainy, sunny, and cloudy.    

We performed a thorough evaluation of the proposed dataset demonstrating important advantages it brings to the state-of-the-art in HDR image reconstruction including: 1) Using the GTA-HDR dataset in combination with other real and synthetic datasets enables significant improvements in the quality of the reconstructed HDR images; and 2) The GTA-HDR dataset fills a gap not covered by any of the publically available real and synthetic datasets and as such, contributes towards better generalization capabilities for HDR image reconstruction.
The main contributions of this work can be summarized as follows:

%The proposed GTA-HDR dataset provides a diverse and generalized set of scenes which is an important advantage to any DL model.
%We perform a set of experiments with the newly created GTA-HDR data and baseline models to find some interesting outcomes - a) We see significant improvement in the quality of generated HDRs while using GTA-HDR dataset along with other real and synthetic datasets either by mixed training or by performing finetuning transfer learning.
%Moreover, models trained on synthetic data in GTA-HDR perform convincingly better when tested on real datasets,
%b) GTA-HDR generalizes a DL model better than any other existing datasets,
%c) GTA-HDR with LDRs having multiple exposures and contrast levels generalize big models significantly well,
%d) GTA-HDR being a completely synthetic dataset fills the gap in the feature domain which is left void by the existing real or mixed datasets hence contributing towards the better generalizing capability of DL models,
%e) Histogram-equalized LDRs play a significant role in the improvement of CNN-based model performance (in the case of any dataset),
%and f) Data and feature fusion as well as self-attention mechanism further improves model performances convincingly.

\begin{itemize}
\item Proposing GTA-HDR, a novel large-scale synthetic dataset and data collection pipeline to complement existing real and synthetic datasets for HDR image reconstruction (see Section~\ref{sec:dataset}).
\item Performing thorough experimental validation using existing real and synthetic datasets and state-of-the-art methods to highlight the contribution of the proposed dataset to the quality of HDR image reconstruction and recovery of image details with high fidelity (see Section~\ref{sec:results}).
\item Demonstrating the contribution of the proposed dataset by illustrating its impact on the state-of-the-art in other computer vision tasks including 3D human pose estimation, human body part segmentation, and holistic scene segmentation (see Section~\ref{sec:results}).
\end{itemize}

\begin{table*}[t]
\centering
\caption{\textbf{Publicly available datasets for HDR image reconstruction.} See Section~\ref{sec:dataset_characteristics} for description of \emph{In-the-wild, Scene diversity} and \emph{Image diversity}; \emph{GT}: Ground truth; \emph{Dis}: Distorted; \emph{*}: Minimum image resolution.}
\label{tab:datasets}
\begin{tabular}{l|c|c|c|c|c|c|c|c}
\toprule[0.5mm]
\textbf{Dataset} & \textbf{Year} & \textbf{Type} & \textbf{\#HDR\textsubscript{GT}} & \textbf{Resolution} & \textbf{In-the-wild} & \textbf{HDR\textsubscript{Dis}} & \textbf{Scene diversity} & \textbf{Image diversity} \\
\midrule[0.25mm]
HDR-Eye~\cite{nemoto2015visual} & 2015 & Synthetic & 46 & $512\times512$ & \xmark & \xmark & \xmark & \xmark \\
City Scene~\cite{zhang2017learning,hdrsky} & 2017 & Mixed & 41222 & $128\times64$ & \xmark & \xmark & \checkmark & \xmark \\
Kalantari \etal~\cite{kalantari2017deep} & 2017 & Real & 89 & $1500\times1000$ & \xmark & \xmark & \xmark & \xmark \\
Endo \etal~\cite{endo2017deep} & 2017 & Synthetic & 1043 & $512\times512$ & \xmark & \xmark & \xmark & \xmark \\
Eilertsen \etal~\cite{eilertsen2017hdr} & 2017 & Synthetic & 96 & $1024\times768$ & \xmark & \xmark & \xmark & \xmark \\
Lee \etal~\cite{lee2018deep} & 2018 & Synthetic & 96 & $512\times512$ & \xmark & \xmark & \xmark & \xmark \\
Cai \etal~\cite{cai2018learning} & 2018 & Synthetic & 4413 &  $3072\times1620^*$  & \xmark & \xmark & \xmark & \xmark \\
Prabhakar \etal~\cite{prabhakar2019fast} & 2019 & Real & 582 &  $1200\times900^*$  & \xmark & \xmark & \xmark & \xmark \\
LDR-HDR Pair~\cite{jang2020dynamic} & 2020 & Real & 176 & $1024\times1024$ & \xmark & \xmark & \xmark & \xmark \\
HDR-Synth \& HDR-Real~\cite{liu2020single} & 2020 & Mixed & 20537 & $512\times512$ & \xmark & \xmark & \xmark & \checkmark \\
SI-HDR~\cite{hanji2022comparison,hanji2022si} & 2022 & Real & 181 & $1920\times1280$ & \xmark & \xmark & \checkmark & \xmark \\
\midrule[0.25mm]
GTA-HDR (ours) & 2024 & Synthetic & 40000 & \makecell{$512\times512$ \\ $1024\times1024$} & \checkmark & \checkmark & \checkmark & \checkmark \\
\bottomrule[0.5mm]
\end{tabular}
\end{table*}

\section{Related Work}
\label{sec:related_work}
This section provides an overview of previous research on traditional (non-learning) techniques, data-driven learning-based methods, and datasets for HDR image reconstruction from single- and multi-exposed LDR images.

\subsection{Inverse Tone Mapping}
\label{sec:inverse_tone_mapping}
Tone mapping~\cite{han2023high}\mycomment{~\cite{han2023high,rana2019deep,rana2018learning,ak2022rv,alotaibi2023quality,han2023highnew}} is the process of mapping the colors of HDR images capturing real-world scenes with a wide range of illumination levels to LDR images appropriate for standard displays with limited dynamic range.
Inverse tone mapping~\cite{wang2021deep} is the reverse process accomplished with either traditional (non-learning) methods or data-driven learning-based approaches.
%Fig.~\ref{fig:tone_mapping} illustrates an overview of the tone mapping pipeline and the process of inverse tone mapping using a data-driven learning-based model.
Given the sensor irradiance $E$ and the exposure time $\Delta t$, the function $f_{crf}(E\Delta t)$ represents the tone mapping process, which outputs $I_{LDR}$ images given $I_{HDR}$ images captured by the camera sensor.
The main goal of any HDR image reconstruction technique is to reverse the tone mapping process using another function $f_{crf}^{-1}(I_{LDR})/\Delta t$, which outputs reconstructed $\hat{I_{HDR}}$ images given $I_{LDR}$ images.
The main challenge is that the steps in $f_{crf}(E\Delta t)$ are generally not reversible~\cite{le2023single}.
%Hence, to invert the non-linearities of this function, we need learning-based methods.
We can, however, approximate the reverse process with a data-driven learning-based model $f_{DL}(I_{LDR}, \Theta)$, which reconstructs $\hat{I_{HDR}}$ images given $I_{LDR}$ images, where $\Theta$ denotes the model parameters.
An illustration of the tone mapping pipeline and the process of inverse tone mapping \mycomment{using a data-driven learning-based model }is provided in the Supplementary Material.
%Good generalization capabilities of the learning-based model require a significant amount of representative and diverse data for optimizing the model parameters $\theta$.

\subsubsection{Non-Learning Methods}
Luzardo \etal~\cite{luzardo2018fully} described an inverse tone mapping operator that allows higher peak brightness (\ie., over 1000 nits) while converting LDR images to HDR counterparts.
The process helps preserve the artistic intent of the reconstructed HDR images.
Kovaleski and Oliveira~\cite{kovaleski2014high} focused on enhancing the over- and under-exposed regions of images using cross-bilateral filtering.
Huo \etal~\cite{huo2014physiological} presented an inverse tone mapping technique based on the human visual system.
The approach uses human retina response to model the inverse local retina response using local luminance adaptation in the image.
Masia \etal~\cite{masia2017dynamic} tried to address the ill-exposed areas of input LDR images, which are more prone to generate artifacts.
This method uses an automatic global reverse tone mapping operator based on gamma expansion along with automatic parameter calculation based on image statistics.
Bist~\etal~\cite{bist2017tone} proposed a gamma correction-based approach that adapts to the target lighting styles of the images.
This work also added a color correction-based operator that reconstructs the intended colors in the HDR image.

\subsubsection{Learning-Based Methods}
Khan \etal~\cite{khan2019fhdr} proposed a feedback mechanism based on a Convolutional Neural Network to generate HDR images from single-exposed LDR inputs.
Barua \etal~\cite{barua2023arthdr} utilized multi-exposed features and perceptual losses along with low- and mid-level feature guidance in generating visually accurate HDR images.
Le \etal~\cite{le2023single} leveraged a Neural Network-based camera response inversion architecture to generate pixel radiance and hallucination details for various exposures.
Liu \etal~\cite{liu2020single} proposed an architecture consisting of three Convolutional Neural Networks that approximates the three sub-tasks in the tone mapping process but in the reverse order.
Li and Fang \etal~\cite{li2019hdrnet} presented a combination of an attention mechanism and a Convolutional Neural Network that can recover over- and under-exposed regions of LDR images.
%and also look into the problems of color quantization.
Santos \etal~\cite{santos2020single} leveraged a feature masking mechanism that helps in reconstructing saturated pixels of LDR images resulting in better visual and perceptual quality of the reconstructed HDR images.
Eilertsen \etal~\cite{eilertsen2017hdr} proposed a Convolutional Neural Network for accurate prediction of HDR pixels from the complex under- and over-exposed counterparts in input LDR images.
Luzardo \etal~\cite{luzardo2020fully} tried to overcome the low peak brightness issues in the reconstructed HDR images to enhance the artistic intent.
Cao \etal~\cite{cao2023decoupled} proposed a method that combines the outputs of preliminary HDR results from a channel-decoupled kernel and pixel-wise output from another architecture resulting in high-quality HDR images.
Jang \etal~\cite{jang2020dynamic} explored the concept of the histogram and color differences between HDR and LDR pairs of multiple exposures to reconstruct HDR images.

Recently Neural Radiance Fields~\cite{mildenhall2021nerf} were used to learn implicit color and radiance fields and perform photo-realistic HDR view synthesis~\cite{huang2022hdr}\mycomment{~\cite{mildenhall2022nerf, huang2022hdr}}.
The literature also provides examples of proposed video-based approaches~\cite{yang2023learning}\mycomment{~\cite{yang2023learning,anand2021hdrvideo,chen2021new,khan2022deephs}} using various learning-based methods including Generative Adversarial Networks and Convolutional Neural Networks.
Finally, some methods primarily deal with dynamic scenes~\cite{kalantari2017deep}\mycomment{~\cite{kalantari2017deep,wu2018deep,pu2020robust,gallo2016stack}}, while others address multiple tasks, \eg, denoising, deblurring, super-resolution, and demosaicing~\cite{mildenhall2022nerf}\mycomment{~\cite{mildenhall2022nerf, li2019feedback,johson2016perceptual}}, and fuse them into an HDR image reconstruction pipeline.

\begin{figure*}[t]
\centering
\includegraphics[width=\linewidth]{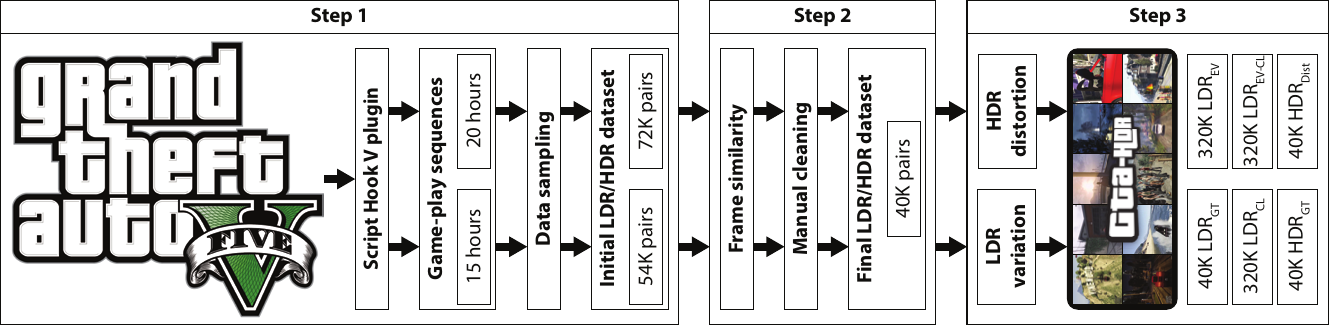}
\caption{\textbf{GTA-HDR dataset collection pipeline.} See Section~\ref{sec:dataset_collection} for detailed description of the three steps; \emph{GT}: Ground truth; \emph{Dis}: Distorted; \emph{EV}: Exposure value; \emph{CL}: Contrast level. \textit{Note}: The GTA-V logo is retrieved from Google Images.}
\label{fig:dataset_collection_pipeline}
\end{figure*}

\subsection{Datasets}
\label{sec:datasets}
In the last several years there has been a shift in the research on HDR image reconstruction from traditional (non-learning) methods to data-driven learning-based approaches based on Generative Adversarial Networks~\cite{raipurkar2021hdr}, Convolutional Neural Networks~\cite{shin2018cnn}, Diffusion Models~\cite{dalal2023single}\mycomment{~\cite{croitoru2023diffusion,dalal2023single}}, and Neural Radiance Fields~\cite{huang2022hdr}\mycomment{~\cite{mildenhall2021nerf,huang2022hdr}}.
%This is due to the success of DL models in various computer vision tasks.
These data-driven techniques require a significant amount of training data drawing attention to the limitations of the current publicly available datasets.
Table~\ref{tab:datasets} summarizes existing public datasets.
%for the task of HDR image reconstruction.
%The datasets can be categorized broadly into real and synthetic.

\subsubsection{Real Datasets}
Real datasets~\cite{sen2012robust,tursun2016objective, dang2015raise,kalantari2017deep,prabhakar2019fast,jang2020dynamic, hanji2022comparison, hanji2022si} include images with sufficient resolution, however, their main limitation is their size (\ie, low number of images). %lack of sufficient amount of ground truth HDR images).
It is difficult, time-consuming, and costly to collect real-world data using an HDR camera that covers a variety of scenes (\eg, indoor, outdoor, in-the-wild), lightning conditions, and image characteristics (\eg, different levels of contrast, radiance, saturation). The RAISE dataset~\cite{dang2015raise} consists of real images and is of moderate size but it lacks ground truth HDR images and therefore is applicable for evaluation purposes only.

\subsubsection{Synthetic Datasets}
Synthetic datasets~\cite{nemoto2015visual,endo2017deep,eilertsen2017hdr,lee2018deep,cai2018learning}, on the other hand, provide ground truth HDR images, however, only a few consist of large amounts of images.
In addition, these datasets generally lack images with appropriate resolution and diversity.
Although the synthetic video dataset proposed in~\cite{kim2019deep} is originally designed for the development of super-resolution video generation methods, it can potentially %The synthetic video dataset proposed in~\cite{kim2019deep} might 
support the development of data-driven learning-based HDR image reconstruction methods after appropriate data pre-processing.

\subsubsection{Mixed Datasets}
Several datasets~\cite{zhang2017learning,hdrsky,liu2020single} include images from both real and synthetic scenes.
These datasets provide a sufficient amount of ground truth HDR images, however, they lack appropriate image resolution and diversity.

\subsubsection{Proposed Dataset}
The proposed GTA-HDR dataset addresses the identified gaps in the current publicly available datasets for HDR image reconstruction.
Specifically:  
1) GTA-HDR is a large-scale (\ie, $40$K ground truth HDR images) synthetic dataset sampled from
the GTA-V video game data which utilizes ray tracing~\cite{glassner1989introduction} technology to simulate the physics behind light and shadows;
2) GTA-HDR includes HDR images with sufficient resolution (\ie., $512 \times 512$ and $1024 \times 1024$), and;
3) GTA-HDR includes HDR images capturing a diverse set of scenes including different locations (\eg, indoor, urban, rural, in-the-wild), different lighting conditions (\eg, morning, midday, evening, night), and different weather and season conditions (\eg, summer, winter, snowy, rainy, sunny, cloudy).

\subsection{No-Reference Quality Assessment}
Due to the lack of ground truth HDR images and high costs of resource-demanding full-reference quality metrics~\cite{alotaibi2023quality} such as High Dynamic Range Visual Differences Predictor~\cite{narwaria2015hdr}\mycomment{~\cite{mantiuk2011hdr,narwaria2015hdr}}, research has also accelerated in the field of no-reference HDR image quality assessment~\cite{banterle2023nor,yan2019naturalness}\mycomment{~\cite{hu2021toward,banterle2020nor,banterle2023nor,artusi2019efficient}}.
Several approaches have tried to approximate the High Dynamic Range Visual Differences Predictor score for perceptual quality assessment of reconstructed HDR images using data-driven learning-based methods, such as Convolutional Neural Networks~\cite{banterle2020nor,banterle2023nor,artusi2019efficient}.
Other approaches have tried to approximate general quality score metrics of images (HDR or LDR), such as Peak Signal-to-Noise Ratio and Structural Similarity Index Measure, using Convolutional Neural Networks and distortion maps~\cite{ravuri2019deep}\mycomment{~\cite{ravuri2019deep,yan2019naturalness}}.
%using deep Convolutional Neural Networks and distortion maps (similar to full reference metrics such as PSNR and SSIM).

The GTA-HDR dataset also contributes a set of distorted HDR images along with the ground truth HDR and LDR images, which addresses the limitations of current datasets concerning no-reference image quality assessment for both HDR and tone-mapped HDR and LDR images.

\begin{figure*}[t]
\centering
\includegraphics[width=\linewidth]{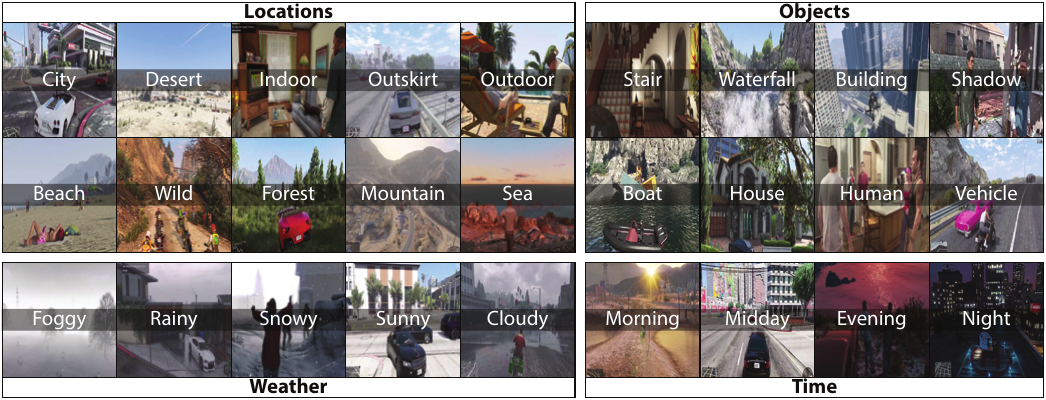}
\caption{\textbf{GTA-HDR dataset scene diversity.} Samples from the GTA-HDR dataset with multiple variations in location, weather, objects and time. The scene diversity ensures a thorough coverage of pixel colors, brightness, and luminance.}
\label{fig:scene_diversity}
\end{figure*}

\section{GTA-HDR Dataset}
\label{sec:dataset}
The GTA-HDR dataset addresses some of the limitations of the existing datasets for HDR image reconstruction.
%Our dataset fills the research gap (see Fig. \ref{fig:supervised_learning_pipelines}) as described in sections \ref{sec:introduction} and \ref{sec:datasets}.
The main characteristics of GTA-HDR are the diversity of scenes and variety of images included in the dataset (\eg, forests, mountains, coasts, cities).
GTA-HDR includes scenes from different times (\eg, morning, evening, day-time, night) and different weather conditions (\eg, rainy, snowy, sunny, misty).
This variety of scenes is an expensive and effort-demanding task to collect in a real-world context.
To our knowledge, this work is the first to use video game data to collect and curate synthetic \{LDR/HDR\} image pairs to support the development of inverse tone mapping data-driven methods.
%Table~\ref{tab:datasets} shows a concise comparison of the existing datasets with GTA-HDR dataset \wrt different features which we explain in the coming sections.

\subsection{Dataset Collection}
\label{sec:dataset_collection}
We performed a thorough data collection and curation, adopting a similar approach as described in~\cite{zhang2021simulation}\mycomment{~\cite{zhang2021simulation,johnson2016driving}}.
We used $2$ full game-play sequences (\ie, playing the story from the beginning until the end) from the GTA-V game to extract \{LDR/HDR\} image pairs at multiple resolutions (\ie., $512 \times 512$ and $1024 \times 1024$).
GTA-V has built-in HDR-10\mycomment{\footnote{http://www.adriancourreges.com/blog/2015/11/02/gta-v-graphics-study/}} support for displaying video sequences on HDR displays.
Fig.~\ref{fig:dataset_collection_pipeline} depicts the entire data collection pipeline for GTA-HDR.

In line with~\cite{zhang2021simulation}\mycomment{~\cite{zhang2021simulation,johnson2016driving}}, we used Script Hook V\footnote{http://www.dev-c.com/gtav/scripthookv/}\mycomment{\footnote{ https://www.gta5-mods.com/tools/scripthookv-net}} plugin to capture HDR frames from the GTA-V game-play sequences.
Other available tools for GTA-V game data extraction include RenderDoc Debugger\footnote{https://github.com/baldurk/renderdoc} and customized RenderDoc for the Game Data platform\footnote{https://github.com/xiaofeng94/renderdoc_for_game_data}.
The normal duration of a GTA-V game-play is approximately $31.5$ hours (\ie., going through all basic aspects of the story), and it can extend up to about $82$ hours (\ie., visit all aspects of the story thoroughly).
We collected data from $2$ game-play sequences, one of around $15$ hours and another of approximately $20$ hours.
During game data extraction, we sampled $1$ frame per second.
This results in approximately $54$K and $72$K \{LDR/HDR\} image pairs, respectively.

The steps followed in the data collection process are as follows (see Fig. \ref{fig:dataset_collection_pipeline}):
1) We used Script Hook V to extract \{LDR/HDR\} image pairs with $1$Hz frequency, which yielded $54$K and $72$K pairs;
2) We removed frames that are similar to the previous or next frames in the sequence to avoid unnecessary increase in dataset size and redundant information.
The similarity\mycomment{\footnote{https://github.com/MKLab-ITI/visil}} between two consecutive frames in the sequence is based on~\cite{kordopatis2019visil}\mycomment{~\cite{kordopatis2019visil,barrow1977parametric}}; we discarded frames that have a similarity score higher than $0.8$.
%determined similarity drawing inspiration from method\footnote{https://github.com/MKLab-ITI/visil} in  which gives us a frame similarity matrix and values.
We also did a manual cleaning of the collected data to remove unwanted scenes (\eg, images containing violence and other objectionable actions or items).
Finally, we ensured that the collected data has an even distribution of scenes from indoor, outdoor, and in-the-wild environments, resulting in a total of $40$K \{LDR/HDR\} pairs;
and 3) We performed transformations on the original LDR images to generate multiple exposure LDR images (\ie., exposure values EV $0$, $\pm1, \pm2, \pm3$, and $\pm4$)~\cite{le2023single} and different contrast levels~\cite{cai2018learning}.
This step results in $40$K $\times$ $25 = 1$M LDR images.
Apart from the $40$K original HDR images, we also generated $40$K distorted HDR images by
utilizing the following state-of-the-art methods:~\cite{khan2019fhdr,le2023single,liu2020single,nguyen2023psenet};
$20$K images were generated using~\cite{khan2019fhdr}, $10$K with~\cite{le2023single}, and~\cite{liu2020single, nguyen2023psenet} were used to produce $5$K images each.
%We stored the final HDR and LDR images in \texttt{.hdr} and \texttt{.png} formats, respectively.

%randomly utilizing one of the following state-of-the-art methods: \cite{khan2019fhdr,le2023single,hanji2022si,li2019hdrnet,santos2020single,eilertsen2017hdr,raipurkar2021hdr,niu2021hdr,endo2017deep,cao2023decoupled,lee2018deep,cai2018learning, hanji2022comparison,liu2020single, nguyen2023psenet}.

\subsection{Dataset Characteristics}
\label{sec:dataset_characteristics}
One of the limitations of existing datasets is the low diversity of scenes and images.
%\emph{Scene diversity} is an important dataset attribute of the proposed GTA-HDR dataset.
%It ensures the variety of collected scenes in the images.
%It also includes the diversity of \emph{In-the-wild} scenes which include captures from forest/mountain/secluded areas (see Figure~\ref{fig:scene_diversity}).
%This is to ensure the dataset is well equipped with comprehensive scene conditions for the learning based methods to generalize the problem well.
%On the other hand, \textit{Image diversity} signifies the variation in exposure and contrast level itself in the images.
%Datasets composed of real images lack scene diversity due to the limitations in cost and effort required for the physical setup of equipment across multiple locations and the manpower it demands.
To address this limitation, the GTA-HDR dataset includes a wide variety of scenes (\eg, indoor, outdoor, in-the-wild, multiple locations, weather conditions, lighting conditions, and time-of-day) and images (\eg, LDR images with $9$ different exposure values EV $0$, $\pm1, \pm2, \pm3$, and $\pm4$ and contrast levels).

\begin{figure*}[t]
\centering
\includegraphics[width=\linewidth]{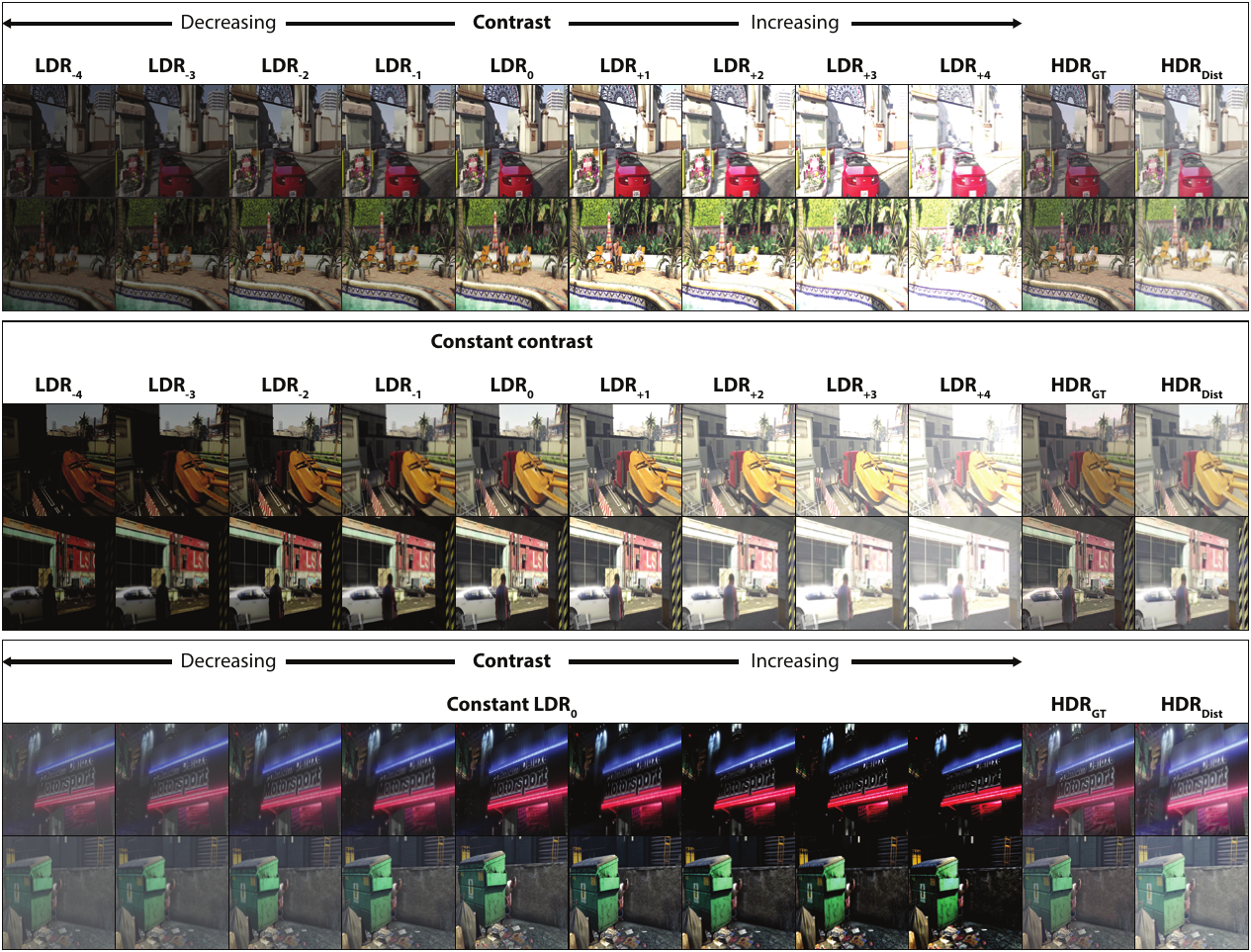}
\caption{\textbf{GTA-HDR dataset image diversity.} Samples from GTA-HDR dataset with multiple exposure values, contrast levels and their combinations. For any image-to-image translation dataset, it is important to include a sufficient samples with diverse range of color hues, saturation, exposure, and contrast levels.} %The final set of images in the dataset amounts to a total of $40$K $\times$ $25 = 1$M LDR, $40$K HDR, and $40$K distorted HDR images.}
\label{fig:image_diversity}
\end{figure*}

\subsubsection{Scene Diversity}
Real-life scenes can have a wide range of variety in terms of locations, landscapes, objects, humans, animals, buildings, weather, and lighting conditions.
%DL models would require samples from all these categories to generalize well in terms of features for learning to convert one image format to another.
%GTA-V game being a good approximation of real-world cities like Los Angeles (USA) consists of most of the above-stated scene categories.
Fig.~\ref{fig:scene_diversity} depicts samples from the GTA-HDR dataset with multiple variations in location, weather, objects and time.
The diverse set of locations ensures a thorough coverage of pixel colors, brightness, and luminance.
The weather conditions contribute to the rich gamut of brightness levels, \eg, sunny weather scenes will have a larger number of bright pixels than cloudy or rainy scenes.
Scenes at different time (\ie, morning, midday, evening, and night) also contribute to different lighting conditions.
The diversity of objects not only captures different color hues but also different contrast and saturation levels.

\subsubsection{Image Diversity}
Images can have a diverse range of color hues, saturation, exposure, and contrast levels.
For any image-to-image translation dataset, it is important to include a sufficient amount of samples from these categories.
Therefore, we introduced different exposure, brightness, and contrast levels in the GTA-HDR dataset.
Considering all variations, the dataset includes $24$ variants of the original LDR images.
The final set of images amounts to a total of $40$K $\times$ $25 = 1$M LDR, $40$K HDR, and $40$K distorted HDR images.
Fig.~\ref{fig:image_diversity} provides samples from the GTA-HDR dataset with $9$ exposure levels (\ie, exposure values EV $0$, $\pm1, \pm2, \pm3$, and $\pm4$) and $9$ contrast levels of the LDR images.
The first two rows show the LDR images with varying contrast and EV levels. 
The LDR images with normal contrast level are shown in the middle, while LDR images with increasing EV and contrast are shown in sequential order toward the right, and vice versa. %on the right and decreasing EV and contrast on the left.
On the extreme right, the corresponding HDR and a sample of distorted HDR (\ie, saturation altered HDR) images are illustrated.
Here, saturation alteration, contrast alteration, color hue alteration, and noise addition are applied to the HDR to produce the respective distorted HDR images.
Similarly, the second and third rows show the LDR images with only varied EV while the contrast is kept constant (\ie, keeping the contrast level of the original LDR image).
Finally, in the fifth and sixth rows, the EV is kept constant (\ie, keeping the EV $0$ of the original LDR image) and the contrast levels are varied.

\subsubsection{No-Reference Quality Assessment}
To address the data gap for no-reference image quality assessment, the GTA-HDR dataset contributes a set of distorted HDR along with the ground truth HDR and LDR images.
The distorted HDR images can be utilized to develop no-reference quality assessment methods, \eg, by adopting a methodology similar to the ones proposed in~\cite{banterle2020nor,banterle2023nor,artusi2019efficient}:
1) Estimate the full-reference quality scores for pairs of ground truth and distorted HDR images using an existing metric such as PSNR, SSIM, HDR-VDP-2/-3, and LPIPS;
2) Develop a data-driven method using the full-reference quality scores and their corresponding distorted HDR images;
and 3) Utilize the developed model to estimate the quality scores of unseen reconstructed HDR images (no-reference quality assessment).
Similarly, one can develop data-driven methods to estimate the quality scores for tone-mapped LDR and HDR images.

\begin{table*}[t]
\centering
\caption{\textbf{Impact of the GTA-HDR dataset on the performance of the state-of-the-art in HDR image reconstruction.} \emph{R}: Real data combines the datasets proposed in~\cite{kalantari2017deep, prabhakar2019fast,jang2020dynamic} and the real images from the datasets proposed in~\cite{zhang2017learning,hdrsky}; \emph{R $\oplus$ S}: This combination contains the mixed datasets (including both real and synthetic data) proposed in~\cite{zhang2017learning,hdrsky} and the real datasets proposed in~\cite{kalantari2017deep, prabhakar2019fast,jang2020dynamic}; \emph{GTA-HDR}: Proposed synthetic dataset; \emph{E2E}: End-to-end training; \emph{FT}: Finetuning of the original pre-trained models. The performance of all methods is evaluated on a separate dataset proposed in~\cite{liu2020single}.}
\label{tab:hdr_reconstruction1}
\begin{tabular}{l|c|c|c|c|c}
\toprule[0.5mm]
\textbf{Method} & \textbf{Configuration} & \textbf{Datasets} & \textbf{PSNR$\uparrow$} & \textbf{SSIM$\uparrow$} & \textbf{Q-score$\uparrow$} \\
\midrule[0.25mm]
HDRCNN~\cite{eilertsen2017hdr} & E2E & R & 19.1 & 0.67 & 59.2 \\
DrTMO~\cite{endo2017deep} & E2E & R & 19.2 & 0.68 & 60.3 \\
FHDR~\cite{khan2019fhdr} & E2E & R & 24.4 & 0.80 & 65.1 \\
SingleHDR~\cite{liu2020single} & E2E & R & 29.1 & 0.81 & 66.2 \\
HDR-GAN~\cite{niu2021hdr} & E2E & R & 36.9 & 0.92 & 65.3 \\
SingleHDR~\cite{le2023single} & E2E & R & 34.7 & 0.91 & 66.9 \\
ArtHDR-Net~\cite{barua2023arthdr} & E2E & R & 35.1 & 0.91 & 67.2 \\
HistoHDR-Net~\cite{barua2024histohdr} & E2E & R & 35.2 & 0.92 & 67.4 \\
\midrule[0.25mm]
HDRCNN~\cite{eilertsen2017hdr} & E2E & R $\oplus$ S & 20.1 & 0.69 & 60.8 \\
DrTMO~\cite{endo2017deep} & E2E & R $\oplus$ S & 20.3 & 0.68 & 61.5 \\
FHDR~\cite{khan2019fhdr} & E2E & R $\oplus$ S & 26.7 & 0.81 & 65.3 \\
SingleHDR~\cite{liu2020single} & E2E & R $\oplus$ S & 30.4 & 0.82 & 66.1 \\
HDR-GAN~\cite{niu2021hdr} & E2E & R $\oplus$ S & 37.8 & 0.94 & 66.7 \\
SingleHDR~\cite{le2023single} & E2E & R $\oplus$ S & 35.2 & 0.92 & 67.1 \\
ArtHDR-Net~\cite{barua2023arthdr} & E2E & R $\oplus$ S & 35.3 & 0.93 & 67.4 \\
HistoHDR-Net~\cite{barua2024histohdr} & E2E & R $\oplus$ S & 35.3 & 0.94 & 67.5 \\
\midrule[0.25mm]
HDRCNN~\cite{eilertsen2017hdr} & E2E / FT & GTA-HDR & 22.4 / 22.1 & 0.72 / 0.71 & 61.3 / 61.4 \\
DrTMO~\cite{endo2017deep} & E2E / FT & GTA-HDR & 23.5 / 23.4 & 0.71 / 0.71 & 64.3 / 64.5 \\
FHDR~\cite{khan2019fhdr} & E2E / FT & GTA-HDR & 27.7 / 27.6 & 0.84 / 0.84 & 68.0 / 68.1 \\
SingleHDR~\cite{liu2020single} & E2E / FT & GTA-HDR & 32.3 / 32.1 & 0.86 / 0.85 & 68.8 / 69.0 \\
HDR-GAN~\cite{niu2021hdr} & E2E / FT & GTA-HDR & 38.7 / 38.5 & 0.94 / 0.93 & 69.5 / 69.7 \\
SingleHDR~\cite{le2023single} & E2E / FT & GTA-HDR & 41.2 / 41.5 & 0.96 / 0.96 & 70.2 / 70.0 \\
ArtHDR-Net~\cite{barua2023arthdr} & E2E / FT & GTA-HDR & 41.6 / 41.5 & 0.97 / 0.97 & 70.4 / 70.2 \\
HistoHDR-Net~\cite{barua2024histohdr} & E2E / FT & GTA-HDR & 41.7 / 41.5 & 0.98 / 0.98 & 70.5 / 70.4 \\
\midrule[0.25mm]
HDRCNN~\cite{eilertsen2017hdr} & E2E / FT & R $\oplus$ S $\oplus$ GTA-HDR & \textbf{22.6} (\textbf{+3.5}) / 22.3 (+3.2) & \textbf{0.70} (\textbf{+0.03}) / 0.69 (+0.02) & 61.6 (+2.4) / \textbf{62.0} (\textbf{+2.8}) \\
DrTMO~\cite{endo2017deep} & E2E / FT & R $\oplus$ S $\oplus$ GTA-HDR &  \textbf{23.6} (\textbf{+4.4}) / 23.5 (+4.3) & 0.71 (+0.03) / \textbf{0.72} (\textbf{+0.04}) & 64.6 (+4.3) / \textbf{64.8} (\textbf{+4.5}) \\
FHDR~\cite{khan2019fhdr} & E2E / FT & R $\oplus$ S $\oplus$ GTA-HDR & \textbf{27.9} (\textbf{+3.5}) / 27.4 (+3.0) & \textbf{0.83 (+0.03)} / \textbf{0.83 (+0.03)} & 67.5 (+2.4) / \textbf{68.1} (\textbf{+3.0}) \\
SingleHDR~\cite{liu2020single} & E2E / FT & R $\oplus$ S $\oplus$ GTA-HDR & \textbf{32.5} (\textbf{+3.4}) / 31.6 (+2.5) & \textbf{0.85} (\textbf{+0.04}) / 0.84 (+0.03) & 68.7 (+2.5) / \textbf{68.8} (\textbf{+2.6}) \\
HDR-GAN~\cite{niu2021hdr} & E2E / FT & R $\oplus$ S $\oplus$ GTA-HDR & \textbf{40.1} (\textbf{+3.2}) / 39.4 (+2.5) & 0.95 (+0.03) / \textbf{0.97} (\textbf{+0.05}) & 69.2 (+3.9) / \textbf{69.5} (\textbf{+4.2}) \\
SingleHDR~\cite{le2023single} & E2E / FT & R $\oplus$ S $\oplus$ GTA-HDR & 41.5 (+6.8) / \textbf{41.9} (\textbf{+7.2}) & 0.97 (+0.06) / \textbf{0.98} (\textbf{+0.07}) & \textbf{70.3} (\textbf{+3.4}) / 70.0 (+3.1) \\
ArtHDR-Net~\cite{barua2023arthdr} & E2E / FT & R $\oplus$ S $\oplus$ GTA-HDR & 41.6 (+6.5) / \textbf{42.1} (\textbf{+7.0}) & \textbf{0.98 (+0.07)} / \textbf{0.98 (+0.07)} & \textbf{71.2} (\textbf{+4.0}) / 70.9 (+3.7) \\
HistoHDR-Net~\cite{barua2024histohdr} & E2E / FT & R $\oplus$ S $\oplus$ GTA-HDR & 41.8 (+6.6) / \textbf{42.3} (\textbf{+7.1}) & \textbf{0.99 (+0.07)} / \textbf{0.99 (+0.07)} & \textbf{71.5} (\textbf{+4.1}) / 71.4 (+4.0) \\
\bottomrule[0.5mm]
\end{tabular}
\end{table*}

\begin{figure}[t]
\centering
\includegraphics[width=0.95\linewidth]{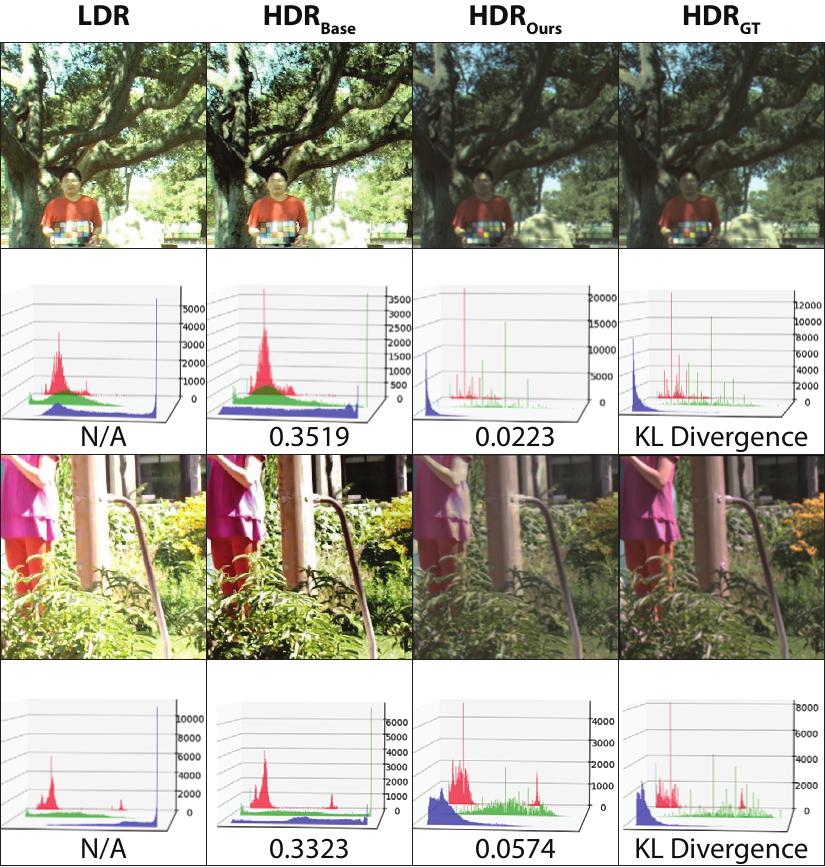}
\caption{\textbf{HDR images reconstructed with and without GTA-HDR as part of the training dataset, along with the RGB histograms and KL-divergence values.} \emph{Base}: HDR images reconstructed with ArtHDR-Net~\cite{barua2023arthdr} trained without GTA-HDR data; \emph{Ours}: HDR images reconstructed with ArtHDR-Net~\cite{barua2023arthdr} trained with GTA-HDR data; \emph{GT}: Ground truth.}
\label{fig:results_all}
\end{figure}

\section{Experiments}
\label{sec:experiments}
We perform a thorough evaluation of the GTA-HDR dataset and assess its contribution to HDR image reconstruction.
% The results demonstrate %from several perspectives 
% the significant benefits brought by the GTA-HDR in the %dataset brings to the 
% task of visually accurate HDR image reconstruction.
We further demonstrate the contribution of the proposed dataset by analyzing its impact on three additional computer vision tasks, including 3D human pose estimation, human body part segmentation, and holistic scene segmentation.
%Additionally, we showcase the usefulness of the dataset (in improving the quality of HDR reconstruction) on other important computer vision tasks. 

\subsection{Experimental Setup}
\label{sec:experimental_setup}
\subsubsection{Implementation}
We used an Ubuntu 20.04.6 LTS workstation with Intel\textregistered~Xeon\textregistered~CPU E5-2687W v3 @ 3.10GHz (20 CPU cores), 126 GB RAM (+ 2 GB swap), NVIDIA GeForce GTX 1080 GPU (8 GB memory), and 1.4 TB SSD.

\subsubsection{Methods}
To evaluate the effectiveness of the GTA-HDR dataset, we utilized existing state-of-the-art HDR image reconstruction methods including FHDR~\cite{khan2019fhdr}\mycomment{(GlabalSIP 2019)}, SingleHDR~\cite{liu2020single,le2023single}\mycomment{(CVPR 2020, WACV 2023)}, HDRCNN~\cite{eilertsen2017hdr}\mycomment{(ACM TOG 2017)}, HDR-GAN~\cite{niu2021hdr}\mycomment{(IEEE TIP 2021)}, DrTMO~\cite{endo2017deep}\mycomment{(ACM TOG 2017)}, ArtHDR-Net~\cite{barua2023arthdr}\mycomment{(APSIPA 2023)}, and HistoHDR-Net~\cite{barua2024histohdr}.
Most of these methods are designed for single-exposed LDR image inputs, while HDR-GAN is designed for three multi-exposed LDR image inputs, SingleHDR~\cite{le2023single} for two or more multi-exposed LDR image inputs without HDR supervision, and ArtHDR-Net attempts to generate perceptually realistic HDR image using features from multi-exposed LDR images. HistoHDR-Net uses histogram-equalized LDR images along with original LDR images as input to facilitate better recovery of color, contrast, saturation, and hue from over/under-exposed regions.

Unless specified otherwise, we used the official implementations and training strategies of the existing state-of-the-art HDR image reconstruction methods. 
Since, SingleHDR~\cite{le2023single} generates multi-exposed LDR images as output, we used the state-of-the-art tool Photomatix~\cite{photomatix} to merge the LDR stack to obtain and HDR image.
For single-exposed LDR image input methods, we consider all of the available LDR images in the datasets.
For multi-exposed LDR image inputs: 1) For methods with three inputs, we consider an overexposed, a normally exposed, and an underexposed LDR image; and 2) For methods with two inputs, we consider an overexposed and an underexposed LDR image.
For existing datasets with single-exposed LDR images, we generated the missing exposure variants.
The diversity of the considered approaches enables a thorough and all-round evaluation of the proposed GTA-HDR dataset.

\subsubsection{Datasets}
\label{sec:dataset_exp}
We considered most %of the 
publicly available datasets for HDR image reconstruction in our experiments~\cite{kalantari2017deep, prabhakar2019fast,jang2020dynamic,zhang2017learning,hdrsky,liu2020single}.
%The combination of the datasets proposed in~\cite{kalantari2017deep, prabhakar2019fast} and the LDR-HDR Pair dataset~\cite{jang2020dynamic} is referred to as real data in the rest of the text.
%In the real dataset scenario, we combined the datasets in~\cite{kalantari2017deep, prabhakar2019fast} and the LDR-HDR pair dataset~\cite{jang2020dynamic}.
%We also consider the City Scene dataset~\cite{zhang2017learning,hdrsky} and HDR-Synth \& HDR-Real dataset~\cite{liu2020single} having both real and synthetic data.
We split the data into a training set consisting of the datasets proposed in~\cite{kalantari2017deep, prabhakar2019fast,jang2020dynamic,zhang2017learning,hdrsky} and test set including the dataset proposed in~\cite{liu2020single}.
We considered different datasets for training and testing to demonstrate that GTA-HDR both enables significant improvements in the quality of the reconstructed HDR images and contributes towards better generalization capabilities of the considered state-of-the-art HDR image reconstruction methods.
%reduce data bias in training and testing processes.
%It also ensures the performance of the GTA-HDR dataset in out-of-domain test data.
%While using splits from the same datasets for training and testing can lead to domain bias in the case of huge datasets, it can be mitigated by ensuring completely separate datasets for test cases.     
Unless specified otherwise, the same protocol for training and testing was used in all experiments to ensure a fair comparison.
All images were resized to $512 \times 512$ resolution before being used for training and testing.
All HDR images displayed in this text have been tone-mapped using the method proposed in~\cite{reinhard2023photographic}.

\subsection{Evaluation Metrics}
\label{sec:evaluation_metrics}
We used three metrics to report the quantitative results.
\textit{High Dynamic Range Visual Differences
Predictor} (HDR-VDP-2)~\cite{mantiuk2011hdr}\mycomment{~\cite{mantiuk2011hdr,narwaria2015hdr}} or Q-Score (Mean Opinion Score Index) is used for evaluation based on the human visual system.
For structural similarity, luminance, and contrast comparisons, the \textit{Structural Similarity Index Measure} (SSIM)~\cite{wang2009mean,wang2004image,wang2004video} is used.
For pixel-to-pixel comparisons, the \textit{Peak Signal-to-Noise Ratio} (PSNR)~\cite{gupta2011modified} is applied.\footnote{Unless specified otherwise, PSNR is computed on tone-mapped original and reconstructed HDR image pairs.}

\section{Results}
\label{sec:results}
\subsection{HDR Image Reconstruction}
\label{sec:hdr_reconstruction}
This section presents the results from state-of-the-art HDR image reconstruction methods trained with different configurations of data, including real data, mixed data, and synthetic data.
%We evaluated the contribution of the  against existing real datasets~\cite{kalantari2017deep, prabhakar2019fast,jang2020dynamic} and mixed datasets~\cite{zhang2017learning,hdrsky,liu2020single} using several state-of-the-art methods for HDR reconstruction.
%We represent the bucket of only real datasets as `R' and synthetic/mixed datasets as `R + S'.
The results are based on two training strategies chosen to evaluate the contributions of the proposed GTA-HDR dataset: 1) End-to-end training (\ie., the models are fully trained with different combinations of data as stated in Section~\ref{sec:dataset_exp}).
%either using only GTA-HDR or a mix of existing datasets, \eg, (R + S) and GTA-HDR in a \hl{blended fashion}),
and 2) Finetuning (\ie., only the final layers of the pre-trained original models are trained with different combinations of data).
%training only the final layers of the pre-trained model using either GTA-HDR alone or a mix of existing datasets and GTA-HDR again in a blended fashion).
The quantitative results of this experiment are summarized in Table~\ref{tab:hdr_reconstruction1}.
%BT strategy signifies the process of training where we mix existing datasets with GTA-HDR and train the selected baseline models.
%FT strategy is used to incorporate transfer learning into the baseline models by considering the pre-trained baseline models and training only the final layers of the model by mixing data with the GTA-HDR dataset while freezing the rest of the model.
%Fig. \ref{fig:finetune} depicts two types of finetuning (FT) strategies we adopt in our work.
%The pink shade depicts the frozen part of the model which we use as it is like a pre-trained model and the grey shade shows the part we finetune using the FT strategy.
The results show a consistent improvement in PSNR, SSIM, and HDR-VDP-2 (Q-score) for all methods, after including the GTA-HDR dataset in the training process.
Furthermore, the results also demonstrate that all the considered state-of-the-art methods trained with GTA-HDR data alone achieved better performance than when they are trained with existing real and synthetic datasets (see third sub-table). 
Moreover, Table~\ref{tab:hdr_reconstruction1} shows how the performance of the state-of-the-art methods improve consistently when we add more variations to the training data, starting with real data in first sub-table and eventually the full mixture of real, synthetic, and GTA-HDR in the last sub-table. 
We also observe that adding variation to data leads to better results from the second sub-table which has a combination of existing real and synthetic datasets.  
%(Table~\ref{tab:hdr_reconstruction1}, first and second sub-tables).(in Table~\ref{tab:hdr_reconstruction1}, third sub-table)
Further improvements are achieved when mixing the existing datasets with GTA-HDR in both end-to-end and fine-tuning strategies.
%(Table~\ref{tab:hdr_reconstruction1}, fourth sub-table).
%We have highlighted the improvement for each of the state-of-the-art methods for both BT and FT strategies in purple color.
%We further bold the entries that are better among the two strategies for each of the baseline methods.
%We can see that PSNR is higher for end-to-end blended training configuration in most of the cases except for SingleHDR~\cite{le2023single} and ArtHDR-Net~\cite{barua2023arthdr}.
%SSIM is almost similar for both end-to-end and fine-tuning configurations with minimum differences.
%We also see that the Q-score is better for fine-tuning strategy for all methods except for SingleHDR~\cite{le2023single} and ArtHDR-Net~\cite{barua2023arthdr}.
It is noteworthy that methods such as FHDR~\cite{khan2019fhdr} trained with GTA-HDR data using both end-to-end and fine-tuning strategies perform better than some of the more complicated newer methods, \eg, SingleHDR~\cite{liu2020single, le2023single} and HDR-GAN~\cite{niu2021hdr} in terms of HDR-VDP-2 (Q-score) and SSIM.
%(Table~\ref{tab:hdr_reconstruction1}).

Fig.~\ref{fig:results_all} illustrates examples of HDR images reconstructed %provides a qualitative comparison of histograms for HDR images reconstructed by the state-of-the-art method proposed in
by training~\cite{barua2023arthdr} %trained 
with the GTA-HDR data in an end-to-end fashion.
%(as listed in the second last row of the fourth sub-table in Table~\ref{tab:hdr_reconstruction1})
The histograms of the ground truth and the reconstructed images are also included. %and histograms for HDR images reconstructed by the same method trained without GTA-HDR data.
We can see that the histograms from the method trained with GTA-HDR data (\ie, HDR\textsubscript{Ours}) are more similar to the histograms of ground truth HDR images (\ie, HDR\textsubscript{GT}) than those from the method trained without GTA-HDR data (\ie, HDR\textsubscript{Base}). We also report the Kullback-Leibler (KL)\mycomment{\footnote{https://en.wikipedia.org/wiki/Kullback\%E2\%80\%93Leibler_divergence}} divergence values for tone-mapped HDR\textsubscript{GT} and tone-mapped HDR\textsubscript{Base} and HDR\textsubscript{Ours} using the RGB intensities.
%Here, all images are tone-mapped to the LDR domain to compute KL-divergence values.
We see the average KL divergence of the RGB histogram intensity distributions are significantly lower for HDR\textsubscript{Ours} compared to HDR\textsubscript{Base}. Further results are provided in the Supplementary Material.

\begin{figure*}[t]
\captionsetup[subfigure]{justification=centering}
\centering
\subfloat[InceptionV3]{\includegraphics[width=0.24\linewidth]{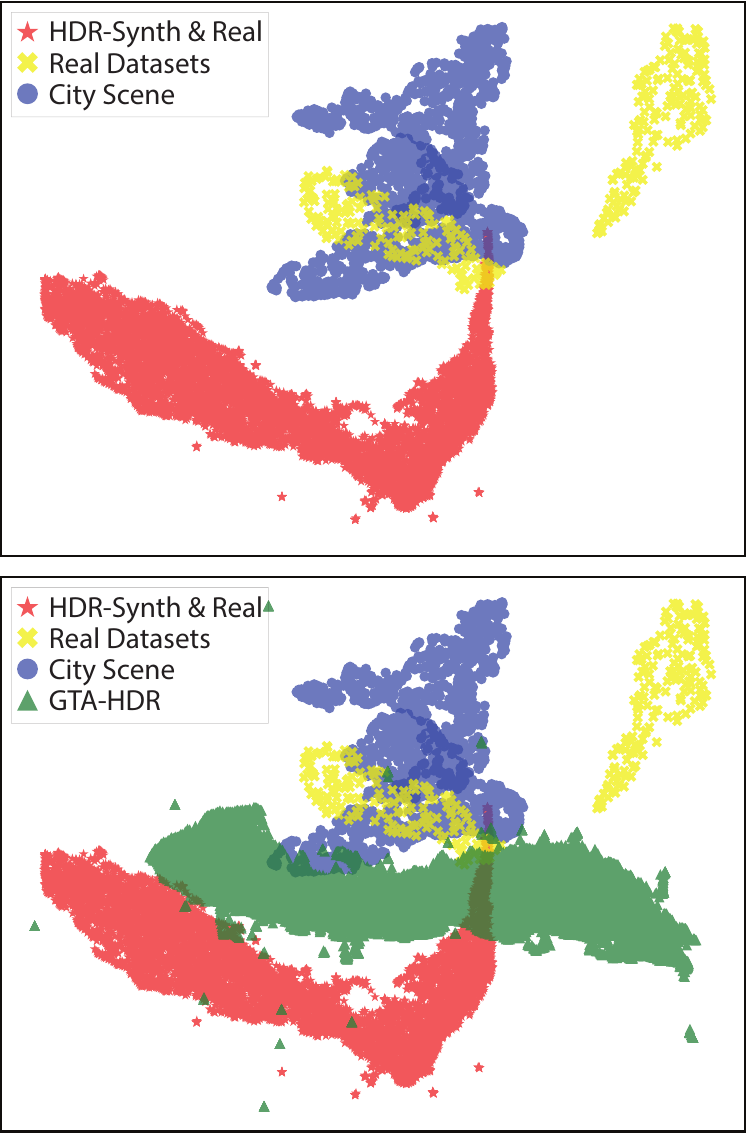}}
\hfill
\subfloat[ResNet50]{\includegraphics[width=0.24\linewidth]{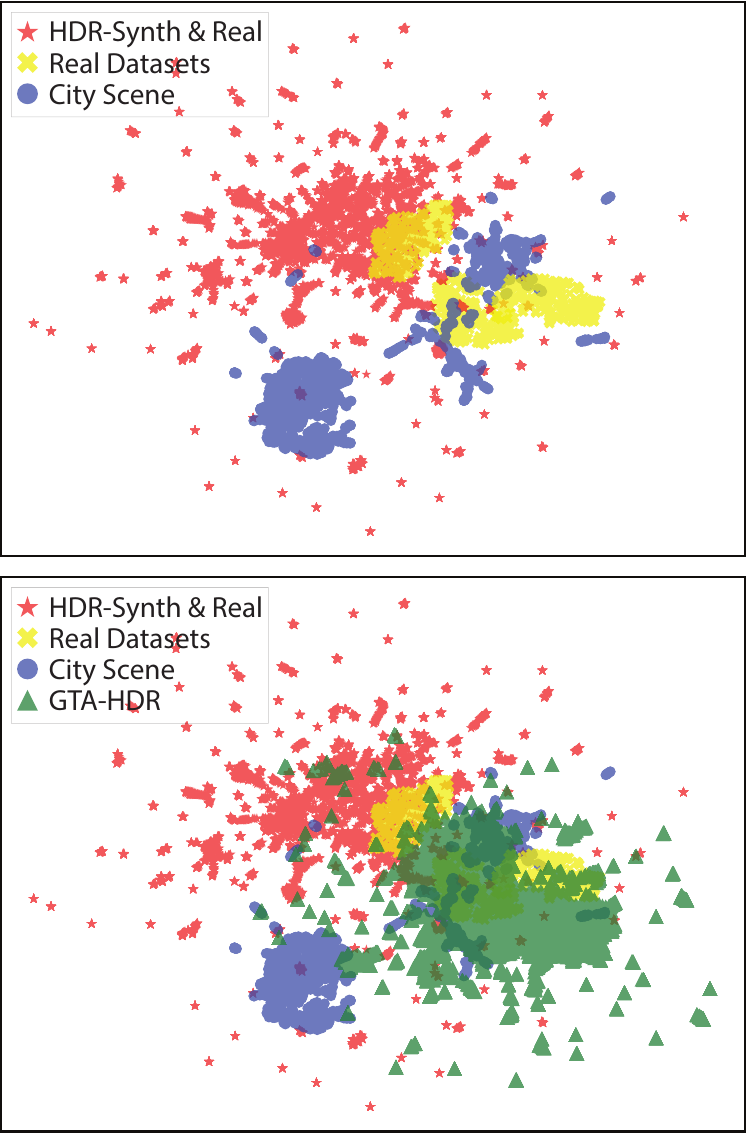}}
\hfill
\subfloat[VGG19]{\includegraphics[width=0.24\linewidth]{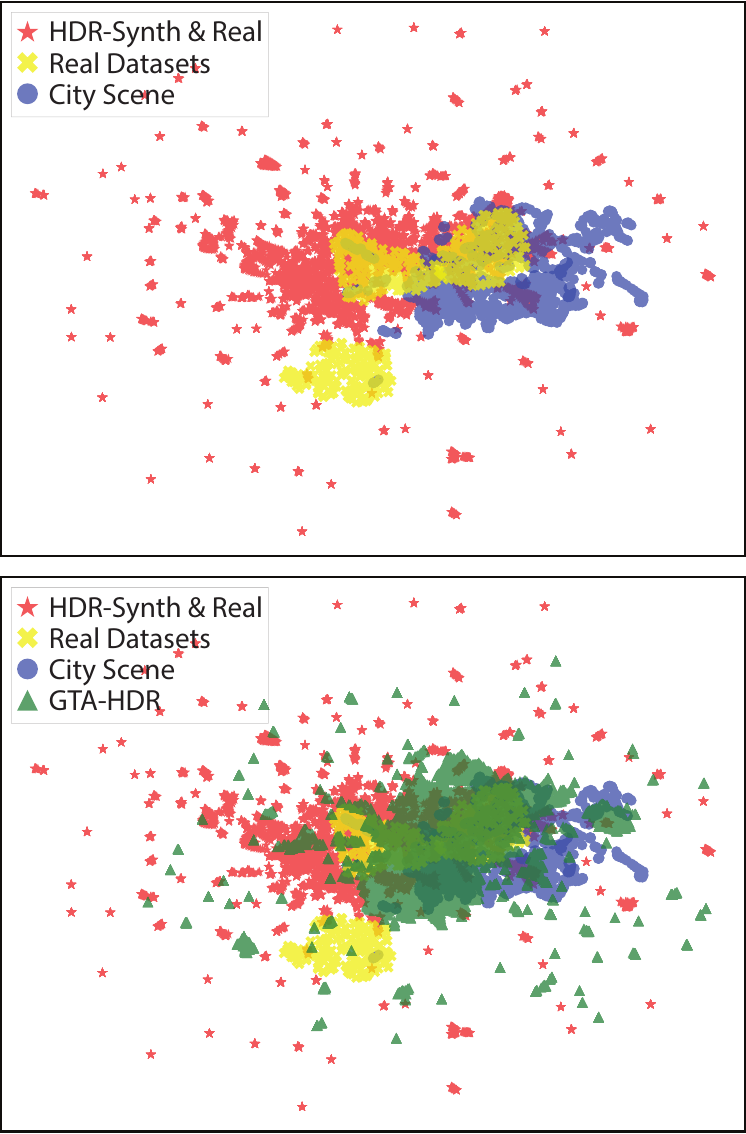}}
\hfill
\subfloat[MobileNet]{\includegraphics[width=0.24\linewidth]{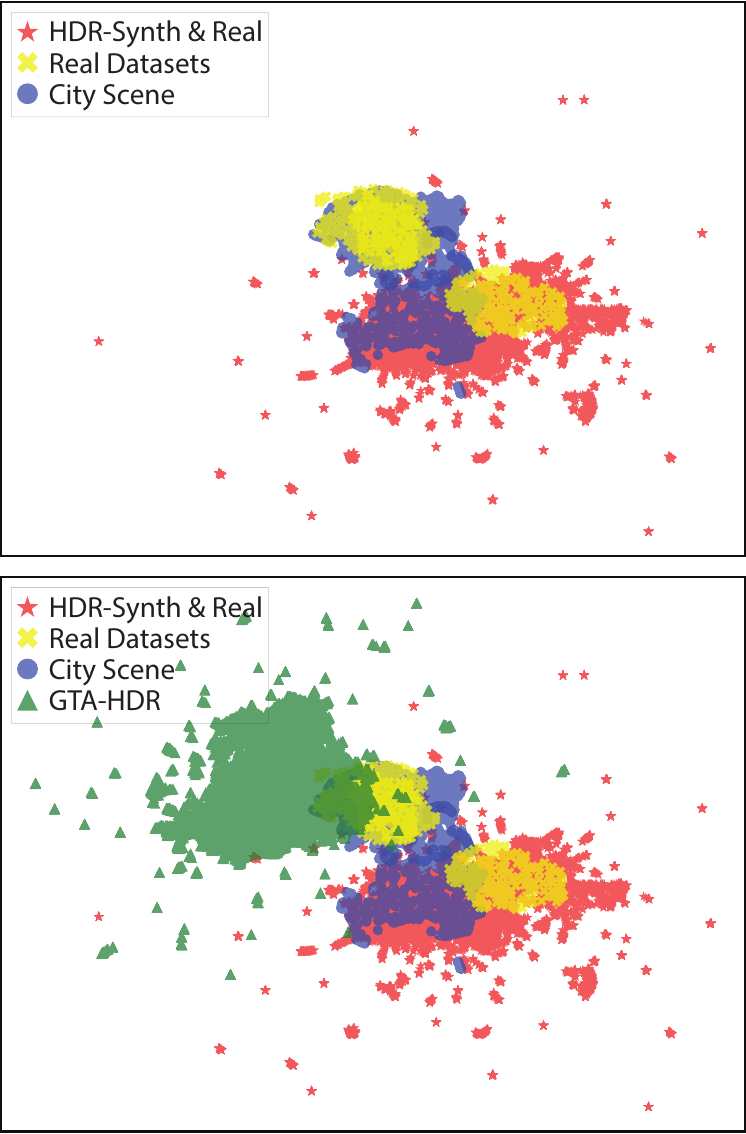}}
\caption{\textbf{Feature space covered by different conventional HDR image reconstruction datasets.} We used UMAP~\cite{mcinnes2018umap} dimension reduction technique to visualize the features extracted from the most common pre-trained feature extraction backbones. \emph{Real Datasets}: Real data combines the datasets proposed in~\cite{kalantari2017deep, prabhakar2019fast,jang2020dynamic}; \emph{City Scene}: Mixed datasets proposed in~\cite{zhang2017learning,hdrsky}; \emph{HDR-Synth \& HDR-Real}: Mixed dataset proposed in~\cite{liu2020single}; \emph{GTA-HDR}: Proposed synthetic dataset.}
\label{fig:features}
\end{figure*}

\begin{table}[t]
\setlength{\tabcolsep}{2.5pt}
\scriptsize
\centering
\caption{\textbf{Performance of SingleHDR~\cite{le2023single}.} Different versions of the state-of-the-art method utilizing different feature extraction backbones, trained with and without GTA-HDR data in an end-to-end fashion. \emph{R $\oplus$ S}: Contains the mixed datasets (including both real and synthetic data) proposed in~\cite{zhang2017learning,hdrsky} and the real datasets proposed in~\cite{kalantari2017deep, prabhakar2019fast,jang2020dynamic}; \emph{GTA-HDR}: Proposed synthetic dataset. All versions of SingleHDR~\cite{le2023single} are evaluated on a separate dataset proposed in~\cite{liu2020single}.}
\label{tab:backbones}
\begin{tabular}{l|c|c|c|c}
\toprule[0.5mm]
\textbf{Method} & \textbf{\#Param} & \textbf{Datasets (training)} & \textbf{PSNR$\uparrow$} & \textbf{SSIM$\uparrow$} \\
\midrule[0.25mm]
SingleHDR (MobileNet) & 13M & R $\oplus$ S & 32.3 & 0.89 \\
SingleHDR (InceptionV3) & 24M & R $\oplus$ S & 32.8 & 0.89 \\
SingleHDR (ResNet50) & 25.6M & R $\oplus$ S & 33.2 & 0.90 \\
SingleHDR (VGG19) & 144M & R $\oplus$ S & 35.2 & 0.92 \\
\midrule[0.25mm]
SingleHDR (MobileNet) & 13M & GTA-HDR & 38.4 & 0.94 \\
SingleHDR (InceptionV3) & 24M & GTA-HDR & 38.8 & 0.95 \\
SingleHDR (ResNet50) & 25.6M & GTA-HDR & 39.5 & 0.95 \\
SingleHDR (VGG19) & 144M & GTA-HDR & 41.2  & 0.96 \\
\midrule[0.25mm]
SingleHDR (MobileNet) & 13M & R $\oplus$ S $\oplus$ GTA-HDR & \textbf{38.6} (\textbf{+6.3}) & \textbf{0.95} (\textbf{+0.06}) \\
SingleHDR (InceptionV3) & 24M & R $\oplus$ S $\oplus$ GTA-HDR & \textbf{39.5} (\textbf{+6.7}) & \textbf{0.95} (\textbf{+0.06}) \\
SingleHDR (ResNet50) & 25.6M & R $\oplus$ S $\oplus$ GTA-HDR & \textbf{40.1} (\textbf{+6.9}) & \textbf{0.96} (\textbf{+0.06}) \\
SingleHDR (VGG19) & 144M & R $\oplus$ S $\oplus$ GTA-HDR & \textbf{41.5} (\textbf{+6.3}) & \textbf{0.97} (\textbf{+0.05}) \\
\bottomrule[0.5mm]
\end{tabular}
\end{table}

\subsection{Scene Diversity}
\label{sec:scene_diversity}
In this experiment, we study the feature space coverage of different datasets as reported by %some of the most 
common feature extraction backbones including MobileNet~\cite{howard2017mobilenets}, InceptionV3~\cite{szegedy2015going}, ResNet50~\cite{shin2018cnn}, and VGG19~\cite{simonyan2014very}.
The results %of the experiment 
reveal a gap in the feature space, \ie, certain regions are not covered by existing real and mixed datasets.
These regions are filled, to a certain extent, by the proposed GTA-HDR dataset.
Features extracted from different backbones can be significantly different based on the underlying architecture, which affects the performance on the downstream tasks (\eg, HDR image reconstruction).
The main goal of feature extraction backbones in LDR to HDR image conversion is the recognition of bright and dark regions and detection of the light source~\cite{shin2018cnn} to ensure that % such that after this process 
underexposed and overexposed regions are %can be 
treated separately. % by the later stages of the model.
%Therefore, it is important to study the most common feature extraction backbones with this perspective in mind.
Fig.~\ref{fig:features} illustrates the feature plots for different datasets.
The first column shows the output of InceptionV3 on existing datasets (top) and the GTA-HDR dataset included (bottom).
There is a significant gap in the feature space between the HDR-Synth \& Real dataset (red) and City Scene dataset (blue) as well as real datasets combined (yellow), which is filled, to some extent, by the GTA-HDR dataset (green).
In the second column (ResNet50), GTA-HDR fills the gap in the lower right corner of the feature space.
Here, we also observed a significant overlap of the GTA-HDR dataset with other datasets.
In the third column (VGG19), the GTA-HDR dataset alone covers a significant area of the feature space which is covered by all other datasets combined.
Finally, in the last column (MobileNet), GTA-HDR extends the feature space by covering a considerable area of the upper left corner along with some overlap with the existing datasets.
%The synthetic data in GTA-HDR is a good candidate for filling the gap left in the feature space by the existing datasets.

%\begin{figure}[t]
%\centering
%\includegraphics[width=\linewidth]{images/feature_plots.pdf}
%\caption{\textbf{\hl{Feature plots.}} We use UMAP~\cite{mcinnes2018umap} dimension reduction to visualize the features extracted from the pre-trained backbones in 2D. Both the axes are principal axes. \textit{R: Real data is a combination of datasets in}~\cite{kalantari2017deep, prabhakar2019fast,jang2020dynamic} (in \textcolor{yellow}{yellow}) and \textit{M: Mixed dataset, \ie, HDR-Synth \& HDR-Real~\cite{zhang2017learning,hdrsky} (in \textcolor{red}{red}), and City Scene~\cite{liu2020single} (in \textcolor{blue}{blue}).} \textit{S: Our proposed synthetic dataset GTA-HDR (in \textcolor{green}{green}).} }
%\label{fig:features}
%\end{figure}

To further investigate the contribution of the GTA-HDR dataset, we replaced the feature extraction block of one of the most recent state-of-the-art methods, SingleHDR~\cite{le2023single} (which originally utilizes VGG19), with the described feature extraction backbones.
%\hl{The method in }\cite{le2023single}\hl{ already uses VGG-19 for its feature extraction purpose, hence we remove the VGG-19 from the base model of }\cite{le2023single}\hl{ and use a simple 3 convolution layers (similar to the one used in the feature block of }\cite{barua2023arthdr})\hl{ to represent the basic model in our experiments.}
Table~\ref{tab:backbones} summarizes the quantitative results of this experiment in terms of PSNR and SSIM.
%\hl{When the model is trained with existing real and mixed datasets in end-to-end fashion, different feature extraction backbones reach better PSNR and SSIM with respect to the basic 3 layer convolutions feature extractor block of SingleHDR}~\cite{le2023single}.
When the model is trained with existing real and mixed datasets in an end-to-end fashion, the observed improvements are proportional to the size of the backbones (\ie., the number of parameters).
However, using only GTA-HDR data for training, there is a significant improvement for all the backbones.
%Finally, when we add the GTA-HDR dataset to the training data we observe a consistent improvement in all cases (see third sub-table of Table \ref{tab:backbones}).
It is worth noting that regardless of the size of the backbone feature extractor, there is an improvement in both PSNR and SSIM when using GTA-HDR.
Interestingly, when including the GTA-HDR dataset, the performance of the considerably smaller (\eg, MobileNet) backbones is better than the large ones (\eg, VGG19) trained without using GTA-HDR.
%This establishes the importance of big synthetic data in small models and big models equally which also establishes the outcome of the plots in Fig. \ref{fig:features}.

\begin{figure*}[t]
\captionsetup[subfigure]{justification=centering}
\centering
\subfloat[Single-exposed]{\includegraphics[width=0.33\linewidth]{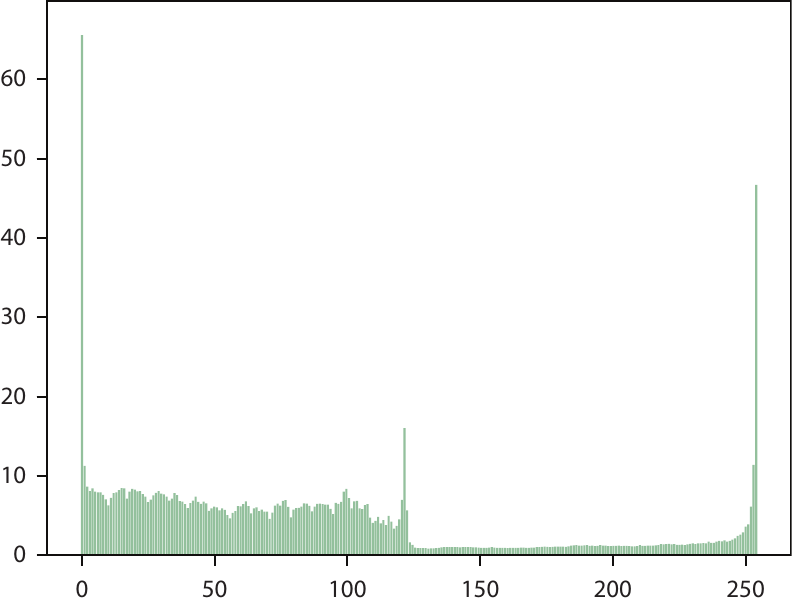}}
\hfill
\subfloat[Multi-exposed]{\includegraphics[width=0.33\linewidth]{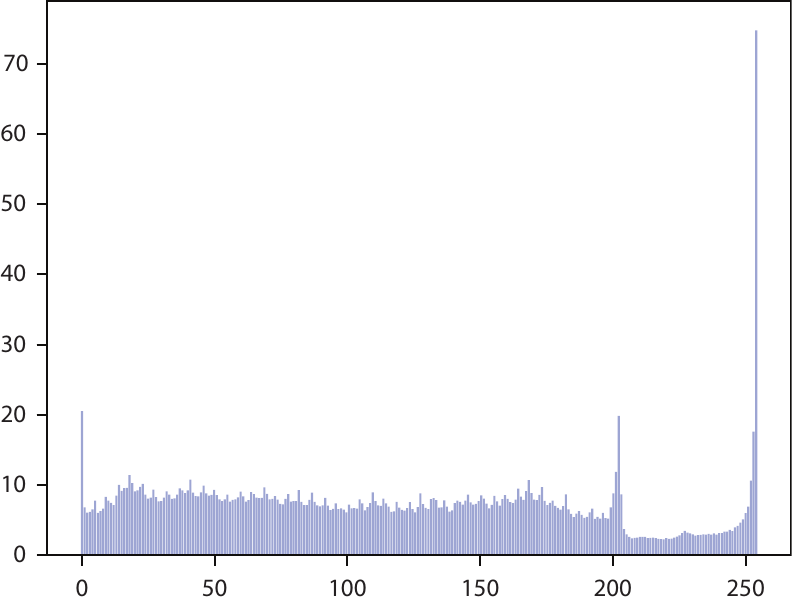}}
\hfill
\subfloat[Multi-exposed and multi-contrast]{\includegraphics[width=0.33\linewidth]{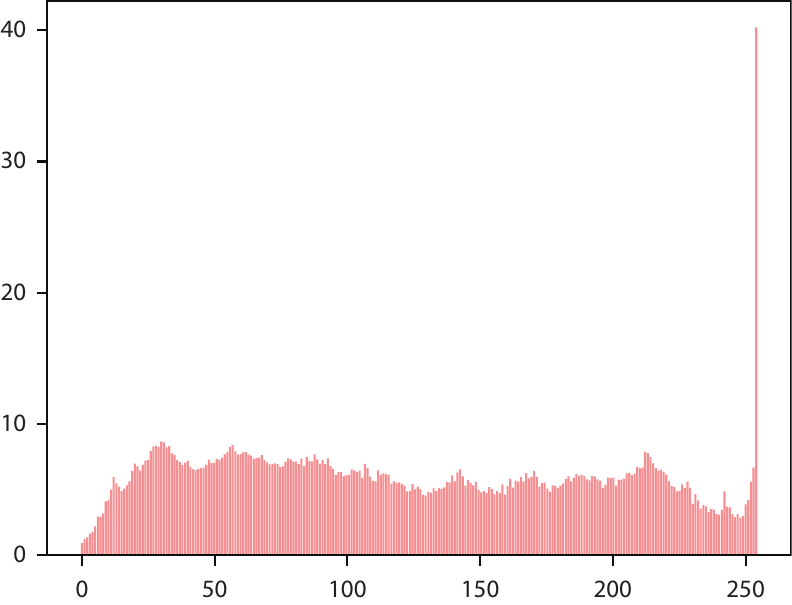}}
\caption{\textbf{Histograms of different variants of the GTA-HDR dataset.} a) GTA-HDR$_{\text{SE}}$ with only single-exposed LDR images, b) GTA-HDR$_{\text{ME}}$ with only multi-exposed LDR images, and c) GTA-HDR$_{\text{FULL}}$ with all LDR images. \label{fig:bias}}
\end{figure*}

\begin{table}[t]
\setlength{\tabcolsep}{2.5pt}
\scriptsize
\centering
\caption{\textbf{Performance of ArtHDR-Net~\cite{barua2023arthdr}.} Assessment of the state-of-the-art method utilizing different training and testing datasets. \emph{SE}: GTA-HDR without exposure and contrast variations; \emph{ME}: GTA-HDR with different exposure levels; \emph{FULL}: Proposed synthetic dataset; \emph{U}: Underexposed LDR images; \emph{O}: Overexposed LDR images; \emph{N}: Normally exposed LDR images. All versions of ArtHDR-Net~\cite{barua2023arthdr} are evaluated on a separate dataset proposed in~\cite{liu2020single}.}
\label{tab:bias}
\begin{tabular}{l|c|c|c|c}
\toprule[0.5mm]
\textbf{Datasets (training)} & \textbf{Datasets (testing)} & \textbf{PSNR$\uparrow$} & \textbf{SSIM$\uparrow$} & \textbf{Q-score$\uparrow$} \\
\midrule[0.25mm]
GTA-HDR\textsubscript{SE} & U & \textbf{42.9} & \textbf{0.99} & \textbf{73.3} \\
GTA-HDR\textsubscript{ME} & U & 40.7 & 0.97 & 71.2 \\
GTA-HDR\textsubscript{FULL} & U & 39.1 & 0.95 & 69.4 \\
\midrule[0.25mm]
GTA-HDR\textsubscript{SE} & O & 32.8 & 0.89 & 64.9 \\
GTA-HDR\textsubscript{ME} & O & 33.9 & 0.92 & 65.7 \\
GTA-HDR\textsubscript{FULL} & O & \textbf{34.7} & \textbf{0.93} & \textbf{66.4} \\
\midrule[0.25mm]
GTA-HDR\textsubscript{SE} & U + O + N & 37.2 & 0.93 & 66.2 \\
GTA-HDR\textsubscript{ME} & U + O + N & 39.7 & 0.95 & 68.5 \\
GTA-HDR\textsubscript{FULL} & U + O + N & \textbf{41.6} & \textbf{0.97} & \textbf{70.4} \\
\bottomrule[0.5mm]
\end{tabular}
\end{table}

%\textcolor{blue}{(GK: How does these experiments mentioned above reflects on scene diversity evaluation? What does scene diversity means? )}
\subsection{Image Diversity}
\label{sec:image_diversity}
%\item Our study establishes the fact that LDRs with diverse exposure and contrast levels can lead to stronger supervision (during training) in DL methods which can in turn improve the accuracy of models (see section \ref{sec:image_diversity}).
%In some cases, to keep the experiments simple and concise we consider only the model FHDR~\cite{khan2019fhdr} (for experiments in tables \ref{tab:bias} to \ref{tab:appl2}) as this model has a unique CNN architecture with a special feedback mechanism similar to the hidden state of a Recurrent Neural Network.
%Here, the HDR reconstruction is guided by multiple iterations of feedback.
The existing datasets include either single-exposed or multi-exposed LDR images and the corresponding HDR images.
The GTA-HDR dataset also introduces LDR images with multi-contrast levels for each HDR image.
%to make the supervision process stronger (see Fig.~\ref{fig:image_diversity}).
In this section, we report the results of an experiment that aims to establish the contribution of multi-exposed and multi-contrast LDR images on the HDR image reconstruction performance.
The experiment includes $3$ versions of the GTA-HDR dataset: 1) GTA-HDR\textsubscript{SE} consists of only single-exposed LDR images; 2) GTA-HDR\textsubscript{ME} comprises only multi-exposed LDR images; and 3) GTA-HDR\textsubscript{FULL} includes all LDR images (\ie., multi-exposed and multi-contrast).
Fig.~\ref{fig:bias} illustrates the histograms of the three versions of the dataset. We treat RGB channels each as intensity values without any conversion. 
The histogram for GTA-HDR$_{\text{SE}}$ reveals % the GTA-HDR dataset with single-exposed LDR images and demonstrates 
that there is a miss-balance in the dataset, where most image pixel intensity values are on the left-hand side of the histogram (\ie., under-exposed pixels $[0, 150]$).
The histogram for GTA-HDR$_{\text{ME}}$ %dataset with multi-exposed LDR images and 
demonstrates that a portion of the miss-balance has been addressed (with a small gap in the interval $[200, 250]$).
The histogram for GTA-HDR$_{\text{FULL}}$ %on the right illustrates the pixel intensity levels distribution after generating multi-contrast levels for each of the LDR images.
confirms that the intensity values %This produces intensity values that are 
are more evenly distributed. % all the intensity levels.

Table~\ref{tab:bias} summarizes the quantitative results of this analysis.
In this experiment, we consider the recent state-of-the-art method ArtHDR-Net~\cite{barua2023arthdr}.
The first three rows report the performance of ArtHDR-Net trained with each version of the GTA-HDR dataset and tested on reconstructing HDR images from underexposed LDR images.
The performance of the model trained with GTA-HDR\textsubscript{SE} dataset is surprisingly good in the case of underexposed LDR images, supporting the qualitative results from the left histogram in Fig.~\ref{fig:bias}.
%However, when different exposure and contrast levels are included during training (\ie., GTA-HDR\textsubscript{ME} and GTA-HDR\textsubscript{FULL}) the performance on reconstructing from underexposed LDR drops.
%A further drop in accuracy is seen when we include the contrast levels as well \ie., GTA-HDR\textsuperscript{FULL}.
%This result supports the observation made for validates the point that the baseline GTA-HDR dataset without varied exposure and contrast levels is biased towards the under-exposed LDRs (Fig.~\ref{fig:bias}).
The middle three rows report the performance of ArtHDR-Net trained with each version of the GTA-HDR dataset and tested on reconstructing HDR images from overexposed LDR images.
In this case, the performance steadily increases with the addition of exposure and contrast levels in the training set.
%In case we test on overexposed images only, the model is generalizing better and better with the inclusion of the exposure and contrast levels.
A similar trend can be observed on the combined over- and underexposed LDR images, reported in the bottom three rows.
%Although the baseline GTA-HDR is performing very well in the case of underexposed LDRs, the model is actually generalizing well only in case of training performed on the full GTA-HDR dataset \ie., GTA-HDR\textsuperscript{FULL}.
Images of real-life scenes have diverse exposure and contrast levels suggesting a data distribution similar to the proposed GTA-HDR\textsubscript{FULL} dataset.
In addition, multi-exposed and multi-contrast LDR images help mitigate model bias towards certain classes of images.
Further qualitative results are provided in the Supplementary Material.

\subsection{Downstream Applications}
\label{sec:downstream_applications}
%\item We perform experimentation and analysis of the real and synthetic datasets (with selected recent DL methods) on some downstream vision applications to establish the importance of our study and the impact of synthetic datasets on them (see section \ref{sec:downstream_applications}).
%To establish the effectiveness of our dataset and state-of-the-art models (trained on our dataset), we evaluate a few downstream virtual reality and robot vision applications such as holistic semantic segmentation, 3D human shape and pose estimation, and human body part segmentation.
To further demonstrate the contribution of the proposed GTA-HDR dataset, this section illustrates its impact on the state-of-the-art in other computer vision tasks including 3D human pose estimation, human body part segmentation, and holistic scene segmentation. Qualitative results for all three tasks are provided in the Supplementary Material.

\subsubsection{3D Human Pose and Shape Estimation}
%We compare the performances with original over-/under-exposed LDR images versus our generated HDR images.
We used BEV~\cite{sun2022putting} as a state-of-the-art pre-trained 3D human pose and shape estimator from images.
We tested the BEV model on the reconstructed HDR images from several versions of the state-of-the-art method ArtHDR-Net~\cite{barua2023arthdr}.
%and a variant trained with our GTA-HDR dataset (with E2E and FT strategies).
%Fig.~\ref{fig:appl1} illustrates the contribution of GTA-HDR trained ArtHDR-Net using an end-to-end strategy for the task of 3D human pose and shape reconstruction.
%The figure demonstrates that the performance of the BEV 3D pose estimator is better in the case of an ArtHDR-Net reconstructed HDR image (capturing $4$ people in the scene accurately) compared to the original LDR images (both under- and overexposed).
Table~\ref{tab:appl1} reports the impact of the image pre-processing step (utilizing reconstructed HDR images from different versions of ArtHDR-Net) on BEV performance evaluated on the AGORA~\cite{patel2021agora} 3D human pose dataset.
We report two commonly used metrics, \emph{F1 Score} to measure detection accuracy and \emph{Mean Per Joint Position Error} (MPJPE) to measure pose accuracy.
The results demonstrate that the pre-processing step enables a significant increase in the performance of BEV.

\begin{table*}[t]
\scriptsize
\centering
\caption{\textbf{Impact of the GTA-HDR dataset on the performance of the state-of-the-art in 3D human pose and shape estimation, 2D human body part segmentation, and semantic segmentation.} We used ArtHDR-Net~\cite{barua2023arthdr} trained with different datasets for HDR image reconstruction. The resulting HDR images from 1) AGORA~\cite{patel2021agora} 3D human pose dataset were used by BEV~\cite{sun2022putting} for 3D human pose and shape estimation; 2) COCO-DensePose~\cite{guler2018densepose} dataset were used by CDCL~\cite{lin2020cross} for 2D human body part segmentation; and 3) Cityscapes~\cite{cordts2016cityscapes} dataset were used by SAM~\cite{kirillov2023segment} for semantic segmentation. \emph{R $\oplus$ S}: This combination contains the mixed datasets (including both real and synthetic data) proposed in~\cite{zhang2017learning,hdrsky} and the real datasets proposed in~\cite{kalantari2017deep, prabhakar2019fast,jang2020dynamic}; \emph{GTA-HDR}: Proposed synthetic dataset; \emph{None}: Results without HDR image reconstruction.}
\label{tab:appl1}
\begin{tabular}{l|c||c|c||c||c}
\toprule[0.5mm]
\multicolumn{2}{c||}{} & \multicolumn{2}{c||}{\textbf{3D human pose and shape estimation}} & \multicolumn{1}{c||}{\textbf{2D human body part segmentation}} & \multicolumn{1}{c}{\textbf{Semantic segmentation}} \\
\midrule[0.25mm]
\textbf{Pre-processing} & \textbf{Datasets} & \textbf{F1 (detection)$\uparrow$} & \textbf{MPJPE (pose)$\downarrow$} & \textbf{mIOU\%$\uparrow$} & \textbf{mIOU\%$\uparrow$} \\
\midrule[0.25mm]
None & - & 0.57 & 129 & 66.24 & 54.24 \\
ArtHDR-Net & - & 0.58 & 128.7 & 67.12 & 54.27 \\
ArtHDR-Net & R $\oplus$ S & 0.58 & 128.5 & 67.9 & 54.26 \\
ArtHDR-Net & GTA-HDR & 0.61 & 125.4 & 69.55 & 54.29 \\
ArtHDR-Net & R $\oplus$ S $\oplus$ GTA-HDR & \textbf{0.65} & \textbf{121.9} & \textbf{74.71} & \textbf{54.36} \\
\bottomrule[0.5mm]
\end{tabular}
\end{table*}

\subsubsection{Human Body Part Segmentation}
In this experiment, we used CDCL~\cite{lin2020cross}, a state-of-the-art body part segmentation model.
Similar to the previous case, we tested the model on
reconstructed HDR images from several versions of ArtHDR-Net~\cite{barua2023arthdr}.
%Fig.~\ref{fig:appl2} %provides an example of 
%illustrates the contribution of the GTA-HDR dataset to this task.
%The human body part segmentation results are more accurate with the reconstructed HDR images than the over- or under-exposed LDR images.
%For the over-exposed LDR images, one person is completely missed in the second image.
%For the under-exposed LDR images, the output is noisy and erroneous. Finally, the processed HDRs using ArtHDR-Net~\cite{barua2023arthdr} trained with the GTA-HDR dataset deliver the most accurate segments.
Table~\ref{tab:appl1} reports the impact of the HDR reconstruction step on the COCO-DensePose~\cite{guler2018densepose} dataset, which is used for CDCL performance evaluation.
We use the \emph{Mean Intersection
of Union}\mycomment{~\cite{everingham2015pascal}} (mIOU\%), \ie, the mean of all IoUs between
predicted and ground truth masks to measure the accuracy of the predictions.
We report the average accuracy for all the body parts considered in~\cite{lin2020cross}.
The results establish the advantages of using the proposed pre-processing (\ie, HDR reconstruction) step.

\subsubsection{Semantic Segmentation}
Finally, we report an experiment on another vision application, \ie., holistic semantic segmentation of scenes, which is an important task in robotics and human-robot interaction.
We consider a recent state-of-the-art method called SAM~\cite{kirillov2023segment} as a pre-trained holistic scene segmentation model.
%We compare the output of SAM on HDRs generated by our training dataset (GTA-HDR with E2E) and strategy against normal unprocessed LDRs.
%We see that the accuracy of segmentation of complex objects in our HDRs is higher than the unprocessed LDRs.
%Fig.~\ref{fig:appl3} illustrates the impact of the GTA-HDR trained model ArtHDR-Net~\cite{barua2023arthdr}.
%The objects/buildings in the background are not segmented well in the overexposed LDR images.
%Similarly, in the underexposed LDR images, even the near objects are sometimes segmented erroneously.
%However, in the HDR images, these issues have been mitigated to a great extent.
Table~\ref{tab:appl1} reports the improvements in the SAM output using the HDR reconstruction as a pre-processing step with the Cityscapes~\cite{cordts2016cityscapes} dataset. %We use this dataset as this is curated from a driving car scenario \ie, stereo video of street scenes with segmentation masks.
We use \emph{Mean Intersection
of Union} (mIOU\%) as the accuracy measure for segmentation.
The results show steady improvements in the performance of SAM.

\section{Conclusions}
\label{sec:conclusions}
This work proposes GTA-HDR, a novel large-scale synthetic dataset and data collection pipeline to complement existing real and synthetic datasets for HDR image reconstruction.
The thorough experimental validation using existing real and synthetic datasets and state-of-the-art methods highlights the contribution of the proposed dataset, specifically to the quality of HDR image reconstruction and the recovery of image details with high fidelity.
We further demonstrate the contribution of the proposed dataset by illustrating its impact on the state-of-the-art in other computer vision tasks, including 3D human pose estimation, human body part segmentation, and holistic scene segmentation.
The proposed dataset represents an important contribution that will enable the development of advanced techniques for visually accurate HDR image reconstruction.
Preliminary discussion has been presented for developing no-reference quality assessment methods utilising the GTA-HDR dataset.
This is possible direction for future research in addition to creating a video-based HDR reconstruction dataset.

\section*{Copyright}
\label{sec:copyright}
As per the publisher's policy\footnote{Policy on copyrighted material: http://tinyurl.com/pjfoqo5} for the GTA-V game\footnote{Policy on mods: http://tinyurl.com/yc8kq7vn}, it allows the use of the generated game-play data provided it is used for non-commercial purposes and without any spoilers.

%\section*{Appendix}
%\label{sec:appendix}

\bibliographystyle{IEEEtran}
\bibliography{bibliography}

\begin{thebibliography}{10}
\providecommand{\url}[1]{#1}
\csname url@samestyle\endcsname
\providecommand{\newblock}{\relax}
\providecommand{\bibinfo}[2]{#2}
\providecommand{\BIBentrySTDinterwordspacing}{\spaceskip=0pt\relax}
\providecommand{\BIBentryALTinterwordstretchfactor}{4}
\providecommand{\BIBentryALTinterwordspacing}{\spaceskip=\fontdimen2\font plus
\BIBentryALTinterwordstretchfactor\fontdimen3\font minus
  \fontdimen4\font\relax}
\providecommand{\BIBforeignlanguage}[2]{{%
\expandafter\ifx\csname l@#1\endcsname\relax
\typeout{** WARNING: IEEEtran.bst: No hyphenation pattern has been}%
\typeout{** loaded for the language `#1'. Using the pattern for}%
\typeout{** the default language instead.}%
\else
\language=\csname l@#1\endcsname
\fi
#2}}
\providecommand{\BIBdecl}{\relax}
\BIBdecl

\bibitem{artusi2019overview}
A.~Artusi, R.~K. Mantiuk, T.~Richter, P.~Hanhart, P.~Korshunov, M.~Agostinelli,
  A.~Ten, and T.~Ebrahimi, ``{Overview and evaluation of the JPEG XT HDR image
  compression standard},'' \emph{Journal of Real-Time Image Processing},
  vol.~16, pp. 413--428, 2019.

\bibitem{he2022sdrtv}
G.~He, K.~Xu, L.~Xu, C.~Wu, M.~Sun, X.~Wen, and Y.-W. Tai, ``{SDRTV-to-HDRTV
  via Hierarchical Dynamic Context Feature Mapping},'' in \emph{Proceedings of
  the 30th ACM International Conference on Multimedia}, 2022, pp. 2890--2898.

\bibitem{satilmis2023deep}
P.~Satilmis and T.~Bashford-Rogers, ``{Deep Dynamic Cloud Lighting},''
  \emph{arXiv preprint arXiv:2304.09317}, 2023.

\bibitem{huang2022hdr}
X.~Huang, Q.~Zhang, Y.~Feng, H.~Li, X.~Wang, and Q.~Wang, ``{HDR-NeRF: High
  Dynamic Range Neural Radiance Fields},'' in \emph{Proceedings of the IEEE/CVF
  Conference on Computer Vision and Pattern Recognition}, 2022, pp.
  18\,398--18\,408.

\bibitem{nguyen2023psenet}
H.~Nguyen, D.~Tran, K.~Nguyen, and R.~Nguyen, ``{PSENet: Progressive
  Self-Enhancement Network for Unsupervised Extreme-Light Image Enhancement},''
  in \emph{Proceedings of the IEEE/CVF Winter Conference on Applications of
  Computer Vision}, 2023, pp. 1756--1765.

\bibitem{wu2020hdr}
X.~Wu, H.~Zhang, X.~Hu, M.~Shakeri, C.~Fan, and J.~Ting, ``{HDR reconstruction
  based on the polarization camera},'' \emph{IEEE Robotics and Automation
  Letters}, vol.~5, no.~4, pp. 5113--5119, 2020.

\bibitem{wang2021deep}
L.~Wang and K.-J. Yoon, ``{Deep Learning for HDR Imaging: State-of-the-Art and
  Future Trends},'' \emph{IEEE transactions on pattern analysis and machine
  intelligence}, vol.~44, no.~12, pp. 8874--8895, 2021.

\bibitem{tiwari2015review}
G.~Tiwari and P.~Rani, ``{A Review On High-Dynamic-Range Imaging With Its
  Technique},'' \emph{International Journal of Signal Processing, Image
  Processing and Pattern Recognition}, vol.~8, no.~9, pp. 93--100, 2015.

\bibitem{tursun2015state}
O.~T. Tursun, A.~O. Aky{\"u}z, A.~Erdem, and E.~Erdem, ``{The State of the Art
  in HDR Deghosting: A Survey and Evaluation},'' in \emph{Computer Graphics
  Forum}, vol.~34, no.~2.\hskip 1em plus 0.5em minus 0.4em\relax Wiley Online
  Library, 2015, pp. 683--707.

\bibitem{johnson2015high}
A.~K. Johnson, ``{High Dynamic Range Imaging - A Review},'' \emph{Int. J. Image
  Process.(IJIP)}, vol.~9, p. 198, 2015.

\bibitem{wang2022kunet}
H.~Wang, M.~Ye, X.~Zhu, S.~Li, C.~Zhu, and X.~Li, ``{KUNet: Imaging
  Knowledge-Inspired Single HDR Image Reconstruction},'' in \emph{The 31st
  International Joint Conference On Artificial Intelligence (IJCAI/ECAI 22)},
  2022.

\bibitem{kinoshita2019scene}
Y.~Kinoshita and H.~Kiya, ``{Scene segmentation-based luminance adjustment for
  multi-exposure image fusion},'' \emph{IEEE Transactions on Image Processing},
  vol.~28, no.~8, pp. 4101--4116, 2019.

\bibitem{luzardo2018fully}
G.~Luzardo, J.~Aelterman, H.~Luong, W.~Philips, D.~Ochoa, and S.~Rousseaux,
  ``{Fully-Automatic Inverse Tone Mapping Preserving the Content Creator's
  Artistic Intentions},'' in \emph{2018 Picture Coding Symposium (PCS)}.\hskip
  1em plus 0.5em minus 0.4em\relax IEEE, 2018, pp. 199--203.

\bibitem{kovaleski2014high}
R.~P. Kovaleski and M.~M. Oliveira, ``{High-Quality Reverse Tone Mapping for a
  Wide Range of Exposures},'' in \emph{2014 27th SIBGRAPI Conference on
  Graphics, Patterns and Images}.\hskip 1em plus 0.5em minus 0.4em\relax IEEE,
  2014, pp. 49--56.

\bibitem{huo2014physiological}
Y.~Huo, F.~Yang, L.~Dong, and V.~Brost, ``{Physiological inverse tone mapping
  based on retina response},'' \emph{The Visual Computer}, vol.~30, pp.
  507--517, 2014.

\bibitem{masia2017dynamic}
B.~Masia, A.~Serrano, and D.~Gutierrez, ``{Dynamic range expansion based on
  image statistics},'' \emph{Multimedia Tools and Applications}, vol.~76, pp.
  631--648, 2017.

\bibitem{guo2023single}
B.-C. Guo and C.-H. Lin, ``{Single-Image HDR Reconstruction Based on Two-Stage
  GAN Structure},'' in \emph{2023 IEEE International Conference on Image
  Processing (ICIP)}.\hskip 1em plus 0.5em minus 0.4em\relax IEEE, 2023, pp.
  91--95.

\bibitem{dalal2023single}
D.~Dalal, G.~Vashishtha, P.~Singh, and S.~Raman, ``{Single Image LDR to HDR
  Conversion Using Conditional Diffusion},'' in \emph{2023 IEEE International
  Conference on Image Processing (ICIP)}.\hskip 1em plus 0.5em minus
  0.4em\relax IEEE, 2023, pp. 3533--3537.

\bibitem{kalantari2017deep}
N.~K. Kalantari, R.~Ramamoorthi \emph{et~al.}, ``{Deep High Dynamic Range
  Imaging of Dynamic Scenes},'' \emph{ACM Trans. Graph.}, vol.~36, no.~4, pp.
  144--1, 2017.

\bibitem{endo2017deep}
Y.~Endo, Y.~Kanamori, and J.~Mitani, ``{Deep reverse tone mapping},'' \emph{ACM
  Trans. Graph.}, vol.~36, no.~6, pp. 177--1, 2017.

\bibitem{eilertsen2017hdr}
G.~Eilertsen, J.~Kronander, G.~Denes, R.~K. Mantiuk, and J.~Unger, ``{HDR image
  reconstruction from a single exposure using deep CNNs},'' \emph{ACM
  transactions on graphics (TOG)}, vol.~36, no.~6, pp. 1--15, 2017.

\bibitem{nemoto2015visual}
H.~Nemoto, P.~Korshunov, P.~Hanhart, and T.~Ebrahimi, ``{Visual attention in
  LDR and HDR images},'' in \emph{9th International Workshop on Video
  Processing and Quality Metrics for Consumer Electronics (VPQM)}, no. CONF,
  2015.

\bibitem{lee2018deep}
S.~Lee, G.~H. An, and S.-J. Kang, ``{Deep Chain HDRI: Reconstructing a High
  Dynamic Range Image from a Single Low Dynamic Range Image},'' \emph{IEEE
  Access}, vol.~6, pp. 49\,913--49\,924, 2018.

\bibitem{prabhakar2019fast}
K.~R. Prabhakar, R.~Arora, A.~Swaminathan, K.~P. Singh, and R.~V. Babu, ``{A
  Fast, Scalable, and Reliable Deghosting Method for Extreme Exposure
  Fusion},'' in \emph{2019 IEEE International Conference on Computational
  Photography (ICCP)}.\hskip 1em plus 0.5em minus 0.4em\relax IEEE, 2019, pp.
  1--8.

\bibitem{jang2020dynamic}
H.~Jang, K.~Bang, J.~Jang, and D.~Hwang, ``{Dynamic Range Expansion Using
  Cumulative Histogram Learning for High Dynamic Range Image Generation},''
  \emph{IEEE Access}, vol.~8, pp. 38\,554--38\,567, 2020.

\bibitem{zhang2017learning}
J.~Zhang and J.-F. Lalonde, ``{Learning High Dynamic Range from Outdoor
  Panoramas},'' in \emph{Proceedings of the IEEE International Conference on
  Computer Vision}, 2017, pp. 4519--4528.

\bibitem{hdrsky}
L.~et~al., ``{The Laval HDR sky database},'' \url{http://hdrdb.com/}, 2016,
  [Online; accessed 3-July-2023].

\bibitem{liu2020single}
Y.-L. Liu, W.-S. Lai, Y.-S. Chen, Y.-L. Kao, M.-H. Yang, Y.-Y. Chuang, and
  J.-B. Huang, ``{Single-Image HDR Reconstruction by Learning to Reverse the
  Camera Pipeline},'' in \emph{Proceedings of the IEEE/CVF Conference on
  Computer Vision and Pattern Recognition}, 2020, pp. 1651--1660.

\bibitem{cai2018learning}
J.~Cai, S.~Gu, and L.~Zhang, ``{Learning a Deep Single Image Contrast Enhancer
  from Multi-Exposure Images},'' \emph{IEEE Transactions on Image Processing},
  vol.~27, no.~4, pp. 2049--2062, 2018.

\bibitem{kim2019deep}
S.~Y. Kim, J.~Oh, and M.~Kim, ``{Deep SR-ITM: Joint Learning of
  Super-Resolution and Inverse Tone-Mapping for 4K UHD HDR Applications},'' in
  \emph{Proceedings of the IEEE/CVF international conference on computer
  vision}, 2019, pp. 3116--3125.

\bibitem{sen2012robust}
P.~Sen, N.~K. Kalantari, M.~Yaesoubi, S.~Darabi, D.~B. Goldman, and
  E.~Shechtman, ``{Robust Patch-Based HDR Reconstruction of Dynamic Scenes},''
  \emph{ACM Trans. Graph.}, vol.~31, no.~6, pp. 203--1, 2012.

\bibitem{tursun2016objective}
O.~T. Tursun, A.~O. Aky{\"u}z, A.~Erdem, and E.~Erdem, ``{An Objective
  Deghosting Quality Metric for HDR Images},'' in \emph{Computer Graphics
  Forum}, vol.~35, no.~2.\hskip 1em plus 0.5em minus 0.4em\relax Wiley Online
  Library, 2016, pp. 139--152.

\bibitem{dang2015raise}
D.-T. Dang-Nguyen, C.~Pasquini, V.~Conotter, and G.~Boato, ``{RAISE: A Raw
  Images Dataset for Digital Image Forensics},'' in \emph{Proceedings of the
  6th ACM multimedia systems conference}, 2015, pp. 219--224.

\bibitem{banterle2020nor}
F.~Banterle, A.~Artusi, A.~Moreo, and F.~Carrara, ``{Nor-Vdpnet: A No-Reference
  High Dynamic Range Quality Metric Trained On Hdr-Vdp 2},'' in \emph{2020 IEEE
  International Conference on Image Processing (ICIP)}.\hskip 1em plus 0.5em
  minus 0.4em\relax IEEE, 2020, pp. 126--130.

\bibitem{banterle2023nor}
F.~Banterle, A.~Artusi, A.~Moreo, F.~Carrara, and P.~Cignoni, ``{NoR-VDPNet++:
  Real-Time No-Reference Image Quality Metrics},'' \emph{IEEE Access}, vol.~11,
  pp. 34\,544--34\,553, 2023.

\bibitem{artusi2019efficient}
A.~Artusi, F.~Banterle, F.~Carra, and A.~Moreno, ``{Efficient Evaluation of
  Image Quality via Deep-Learning Approximation of Perceptual Metrics},''
  \emph{IEEE Transactions on Image Processing}, vol.~29, pp. 1843--1855, 2019.

\bibitem{zhang2021simulation}
L.~Zhang, A.~Zhu, S.~Zhao, and Y.~Zhou, ``{Simulation of Atmospheric Visibility
  Impairment},'' \emph{IEEE Transactions on Image Processing}, vol.~30, pp.
  8713--8726, 2021.

\bibitem{yang2023synbody}
Z.~Yang, Z.~Cai, H.~Mei, S.~Liu, Z.~Chen, W.~Xiao, Y.~Wei, Z.~Qing, C.~Wei,
  B.~Dai \emph{et~al.}, ``{SynBody: Synthetic Dataset with Layered Human Models
  for 3D Human Perception and Modeling},'' \emph{arXiv preprint
  arXiv:2303.17368}, 2023.

\bibitem{angus2018unlimited}
M.~Angus, M.~ElBalkini, S.~Khan, A.~Harakeh, O.~Andrienko, C.~Reading,
  S.~Waslander, and K.~Czarnecki, ``{Unlimited Road-scene Synthetic Annotation
  (URSA) Dataset},'' in \emph{2018 21st International Conference on Intelligent
  Transportation Systems (ITSC)}.\hskip 1em plus 0.5em minus 0.4em\relax IEEE,
  2018, pp. 985--992.

\bibitem{richter2017playing}
S.~R. Richter, Z.~Hayder, and V.~Koltun, ``{Playing for Benchmarks},'' in
  \emph{Proceedings of the IEEE International Conference on Computer Vision},
  2017, pp. 2213--2222.

\bibitem{fabbri2021motsynth}
M.~Fabbri, G.~Bras{\'o}, G.~Maugeri, O.~Cetintas, R.~Gasparini, A.~O{\v{s}}ep,
  S.~Calderara, L.~Leal-Taix{\'e}, and R.~Cucchiara, ``{MOTSynth: How Can
  Synthetic Data Help Pedestrian Detection and Tracking?}'' in
  \emph{Proceedings of the IEEE/CVF International Conference on Computer
  Vision}, 2021, pp. 10\,849--10\,859.

\bibitem{hu2021sail}
Y.-T. Hu, J.~Wang, R.~A. Yeh, and A.~G. Schwing, ``{SAIL-VOS 3D: A Synthetic
  Dataset and Baselines for Object Detection and 3D Mesh Reconstruction from
  Video Data},'' in \emph{Proceedings of the IEEE/CVF Conference on Computer
  Vision and Pattern Recognition}, 2021, pp. 1418--1428.

\bibitem{krahenbuhl2018free}
P.~Kr{\"a}henb{\"u}hl, ``{Free Supervision From Video Games},'' in
  \emph{Proceedings of the IEEE conference on computer vision and pattern
  recognition}, 2018, pp. 2955--2964.

\bibitem{richter2016playing}
S.~R. Richter, V.~Vineet, S.~Roth, and V.~Koltun, ``{Playing for Data: Ground
  Truth from Computer Games},'' in \emph{Computer Vision--ECCV 2016: 14th
  European Conference, Amsterdam, The Netherlands, October 11-14, 2016,
  Proceedings, Part II 14}.\hskip 1em plus 0.5em minus 0.4em\relax Springer,
  2016, pp. 102--118.

\bibitem{witcher}
C.~P. RED, ``{The Witcher 3: Wild Hunt},''
  \url{https://www.thewitcher.com/in/en/witcher3}, 2016, [Online; accessed
  3-Dec-2023].

\bibitem{xu2017end}
H.~Xu, Y.~Gao, F.~Yu, and T.~Darrell, ``{End-to-end Learning of Driving Models
  from Large-scale Video Datasets},'' in \emph{Proceedings of the IEEE
  conference on computer vision and pattern recognition}, 2017, pp. 2174--2182.

\bibitem{hanji2022comparison}
P.~Hanji, R.~Mantiuk, G.~Eilertsen, S.~Hajisharif, and J.~Unger, ``{Comparison
  of single image HDR reconstruction methods—the caveats of quality
  assessment},'' in \emph{ACM SIGGRAPH 2022 conference proceedings}, 2022, pp.
  1--8.

\bibitem{hanji2022si}
------, ``{SI-HDR-dataset for comparison of single-image high dynamic range
  reconstruction methods},'' \emph{University of Cambridge}, 2022.

\bibitem{han2023high}
X.~Han, I.~R. Khan, and S.~Rahardja, ``{High Dynamic Range Image Tone Mapping:
  Literature review and performance benchmark},'' \emph{Digital Signal
  Processing}, p. 104015, 2023.

\bibitem{le2023single}
P.-H. Le, Q.~Le, R.~Nguyen, and B.-S. Hua, ``{Single-Image HDR Reconstruction
  by Multi-Exposure Generation},'' in \emph{Proceedings of the IEEE/CVF Winter
  Conference on Applications of Computer Vision}, 2023, pp. 4063--4072.

\bibitem{bist2017tone}
C.~Bist, R.~Cozot, G.~Madec, and X.~Ducloux, ``{Tone expansion using lighting
  style aesthetics},'' \emph{Computers \& Graphics}, vol.~62, pp. 77--86, 2017.

\bibitem{khan2019fhdr}
Z.~Khan, M.~Khanna, and S.~Raman, ``{FHDR: HDR Image Reconstruction from a
  Single LDR Image using Feedback Network},'' in \emph{2019 IEEE Global
  Conference on Signal and Information Processing (GlobalSIP)}.\hskip 1em plus
  0.5em minus 0.4em\relax IEEE, 2019, pp. 1--5.

\bibitem{barua2023arthdr}
H.~B. Barua, G.~Krishnasamy, K.~Wong, K.~Stefanov, and A.~Dhall, ``{ArtHDR-Net:
  Perceptually Realistic and Accurate HDR Content Creation},'' in \emph{2023
  Asia Pacific Signal and Information Processing Association Annual Summit and
  Conference (APSIPA ASC)}.\hskip 1em plus 0.5em minus 0.4em\relax IEEE, 2023,
  pp. 806--812.

\bibitem{li2019hdrnet}
J.~Li and P.~Fang, ``{HDRNET: Single-Image-based HDR Reconstruction Using
  Channel Attention CNN},'' in \emph{Proceedings of the 2019 4th International
  Conference on Multimedia Systems and Signal Processing}, 2019, pp. 119--124.

\bibitem{santos2020single}
M.~S. Santos, T.~I. Ren, and N.~K. Kalantari, ``{Single Image HDR
  Reconstruction Using a CNN with Masked Features and Perceptual Loss},''
  \emph{ACM Transactions on Graphics (TOG)}, vol.~39, no.~4, pp. 80--1, 2020.

\bibitem{luzardo2020fully}
G.~Luzardo, J.~Aelterman, H.~Luong, S.~Rousseaux, D.~Ochoa, and W.~Philips,
  ``{Fully-automatic inverse tone mapping algorithm based on dynamic mid-level
  tone mapping},'' \emph{APSIPA Transactions on Signal and Information
  Processing}, vol.~9, p.~e7, 2020.

\bibitem{cao2023decoupled}
G.~Cao, F.~Zhou, K.~Liu, A.~Wang, and L.~Fan, ``{A Decoupled Kernel Prediction
  Network Guided by Soft Mask for Single Image HDR Reconstruction},'' \emph{ACM
  Transactions on Multimedia Computing, Communications and Applications},
  vol.~19, no.~2s, pp. 1--23, 2023.

\bibitem{mildenhall2021nerf}
B.~Mildenhall, P.~P. Srinivasan, M.~Tancik, J.~T. Barron, R.~Ramamoorthi, and
  R.~Ng, ``{NeRF: Representing Scenes as Neural Radiance Fields for View
  Synthesis},'' \emph{Communications of the ACM}, vol.~65, no.~1, pp. 99--106,
  2021.

\bibitem{yang2023learning}
Y.~Yang, J.~Han, J.~Liang, I.~Sato, and B.~Shi, ``{Learning Event Guided High
  Dynamic Range Video Reconstruction},'' in \emph{Proceedings of the IEEE/CVF
  Conference on Computer Vision and Pattern Recognition}, 2023, pp.
  13\,924--13\,934.

\bibitem{mildenhall2022nerf}
B.~Mildenhall, P.~Hedman, R.~Martin-Brualla, P.~P. Srinivasan, and J.~T.
  Barron, ``{NeRF in the Dark: High Dynamic Range View Synthesis from Noisy Raw
  Images},'' in \emph{Proceedings of the IEEE/CVF Conference on Computer Vision
  and Pattern Recognition}, 2022, pp. 16\,190--16\,199.

\bibitem{raipurkar2021hdr}
P.~Raipurkar, R.~Pal, and S.~Raman, ``{HDR-cGAN: Single LDR to HDR Image
  Translation using Conditional GAN},'' in \emph{Proceedings of the Twelfth
  Indian Conference on Computer Vision, Graphics and Image Processing}, 2021,
  pp. 1--9.

\bibitem{shin2018cnn}
S.~Shin, K.~Kong, and W.-J. Song, ``{CNN-based LDR-to-HDR conversion system},''
  in \emph{2018 IEEE International Conference on Consumer Electronics
  (ICCE)}.\hskip 1em plus 0.5em minus 0.4em\relax IEEE, 2018, pp. 1--2.

\bibitem{glassner1989introduction}
A.~S. Glassner, \emph{{An introduction to ray tracing}}.\hskip 1em plus 0.5em
  minus 0.4em\relax Morgan Kaufmann, 1989.

\bibitem{alotaibi2023quality}
T.~Alotaibi, I.~R. Khan, and F.~Bourennani, ``{Quality Assessment of
  Tone-mapped Images Using Fundamental Color and Structural Features},''
  \emph{IEEE Transactions on Multimedia}, 2023.

\bibitem{narwaria2015hdr}
M.~Narwaria, R.~K. Mantiuk, M.~P. Da~Silva, and P.~Le~Callet, ``{HDR-VDP-2.2: A
  calibrated method for objective quality prediction of high dynamic range and
  standard images},'' \emph{Journal of Electronic Imaging}, vol.~24, no.~1, pp.
  010\,501--010\,501, 2015.

\bibitem{yan2019naturalness}
B.~Yan, B.~Bare, and W.~Tan, ``{Naturalness-aware deep no-reference image
  quality assessment},'' \emph{IEEE Transactions on Multimedia}, vol.~21,
  no.~10, pp. 2603--2615, 2019.

\bibitem{ravuri2019deep}
C.~S. Ravuri, R.~Sureddi, S.~V.~R. Dendi, S.~Raman, and S.~S. Channappayya,
  ``{Deep no-reference tone mapped image quality assessment},'' in \emph{2019
  53rd Asilomar Conference on Signals, Systems, and Computers}.\hskip 1em plus
  0.5em minus 0.4em\relax IEEE, 2019, pp. 1906--1910.

\bibitem{kordopatis2019visil}
G.~Kordopatis-Zilos, S.~Papadopoulos, I.~Patras, and I.~Kompatsiaris, ``{ViSiL:
  Fine-grained Spatio-Temporal Video Similarity Learning},'' in
  \emph{Proceedings of the IEEE/CVF International Conference on Computer
  Vision}, 2019.

\bibitem{niu2021hdr}
Y.~Niu, J.~Wu, W.~Liu, W.~Guo, and R.~W. Lau, ``{HDR-GAN: HDR Image
  Reconstruction from Multi-Exposed LDR Images with Large Motions},''
  \emph{IEEE Transactions on Image Processing}, vol.~30, pp. 3885--3896, 2021.

\bibitem{barua2024histohdr}
H.~B. Barua, G.~Krishnasamy, K.~Wong, A.~Dhall, and K.~Stefanov,
  ``{HistoHDR-Net: Histogram Equalization for Single LDR to HDR Image
  Translation},'' \emph{arXiv preprint arXiv:2402.06692}, 2024.

\bibitem{photomatix}
HDRsoft, ``{Photomatix},'' \url{https://www.hdrsoft.com/}, [Online; accessed
  3-Nov-2023].

\bibitem{reinhard2023photographic}
E.~Reinhard, M.~Stark, P.~Shirley, and J.~Ferwerda, ``{Photographic tone
  reproduction for digital images},'' in \emph{Seminal Graphics Papers: Pushing
  the Boundaries, Volume 2}, 2023, pp. 661--670.

\bibitem{mantiuk2011hdr}
R.~Mantiuk, K.~J. Kim, A.~G. Rempel, and W.~Heidrich, ``{HDR-VDP-2: A
  calibrated visual metric for visibility and quality predictions in all
  luminance conditions},'' \emph{ACM Transactions on graphics (TOG)}, vol.~30,
  no.~4, pp. 1--14, 2011.

\bibitem{wang2009mean}
Z.~Wang and A.~C. Bovik, ``{Mean squared error: Love it or leave it? A new look
  at Signal Fidelity Measures},'' \emph{IEEE signal processing magazine},
  vol.~26, no.~1, pp. 98--117, 2009.

\bibitem{wang2004image}
Z.~Wang, A.~C. Bovik, H.~R. Sheikh, and E.~P. Simoncelli, ``{Image Quality
  Assessment: From Error Visibility to Structural Similarity},'' \emph{IEEE
  transactions on image processing}, vol.~13, no.~4, pp. 600--612, 2004.

\bibitem{wang2004video}
Z.~Wang, L.~Lu, and A.~C. Bovik, ``{Video quality assessment based on
  structural distortion measurement},'' \emph{Signal processing: Image
  communication}, vol.~19, no.~2, pp. 121--132, 2004.

\bibitem{gupta2011modified}
P.~Gupta, P.~Srivastava, S.~Bhardwaj, and V.~Bhateja, ``{A modified PSNR metric
  based on HVS for quality assessment of color images},'' in \emph{2011
  International Conference on Communication and Industrial Application}.\hskip
  1em plus 0.5em minus 0.4em\relax IEEE, 2011, pp. 1--4.

\bibitem{mcinnes2018umap}
L.~McInnes, J.~Healy, and J.~Melville, ``{Umap: Uniform manifold approximation
  and projection for dimension reduction},'' \emph{arXiv preprint
  arXiv:1802.03426}, 2018.

\bibitem{howard2017mobilenets}
A.~G. Howard, M.~Zhu, B.~Chen, D.~Kalenichenko, W.~Wang, T.~Weyand,
  M.~Andreetto, and H.~Adam, ``{Mobilenets: Efficient convolutional neural
  networks for mobile vision applications},'' \emph{arXiv preprint
  arXiv:1704.04861}, 2017.

\bibitem{szegedy2015going}
C.~Szegedy, W.~Liu, Y.~Jia, P.~Sermanet, S.~Reed, D.~Anguelov, D.~Erhan,
  V.~Vanhoucke, and A.~Rabinovich, ``{Going deeper with convolutions},'' in
  \emph{Proceedings of the IEEE conference on computer vision and pattern
  recognition}, 2015, pp. 1--9.

\bibitem{simonyan2014very}
K.~Simonyan and A.~Zisserman, ``{Very deep convolutional networks for
  large-scale image recognition},'' \emph{arXiv preprint arXiv:1409.1556},
  2014.

\bibitem{sun2022putting}
Y.~Sun, W.~Liu, Q.~Bao, Y.~Fu, T.~Mei, and M.~J. Black, ``{Putting People in
  their Place: Monocular Regression of 3D People in Depth},'' in
  \emph{Proceedings of the IEEE/CVF Conference on Computer Vision and Pattern
  Recognition}, 2022, pp. 13\,243--13\,252.

\bibitem{patel2021agora}
P.~Patel, C.-H.~P. Huang, J.~Tesch, D.~T. Hoffmann, S.~Tripathi, and M.~J.
  Black, ``{AGORA: Avatars in geography optimized for regression analysis},''
  in \emph{Proceedings of the IEEE/CVF Conference on Computer Vision and
  Pattern Recognition}, 2021, pp. 13\,468--13\,478.

\bibitem{guler2018densepose}
R.~A. G{\"u}ler, N.~Neverova, and I.~Kokkinos, ``{Densepose: Dense human pose
  estimation in the wild},'' in \emph{Proceedings of the IEEE conference on
  computer vision and pattern recognition}, 2018, pp. 7297--7306.

\bibitem{lin2020cross}
K.~Lin, L.~Wang, K.~Luo, Y.~Chen, Z.~Liu, and M.-T. Sun, ``{Cross-domain
  complementary learning using pose for multi-person part segmentation},''
  \emph{IEEE Transactions on Circuits and Systems for Video Technology},
  vol.~31, no.~3, pp. 1066--1078, 2020.

\bibitem{cordts2016cityscapes}
M.~Cordts, M.~Omran, S.~Ramos, T.~Rehfeld, M.~Enzweiler, R.~Benenson,
  U.~Franke, S.~Roth, and B.~Schiele, ``{The Cityscapes Dataset for Semantic
  Urban Scene Understanding},'' in \emph{Proceedings of the IEEE conference on
  computer vision and pattern recognition}, 2016, pp. 3213--3223.

\bibitem{kirillov2023segment}
A.~Kirillov, E.~Mintun, N.~Ravi, H.~Mao, C.~Rolland, L.~Gustafson, T.~Xiao,
  S.~Whitehead, A.~C. Berg, W.-Y. Lo \emph{et~al.}, ``{Segment anything},''
  \emph{arXiv preprint arXiv:2304.02643}, 2023.

\end{thebibliography}


\begin{thebibliography}{10}
\providecommand{\url}[1]{#1}
\csname url@samestyle\endcsname
\providecommand{\newblock}{\relax}
\providecommand{\bibinfo}[2]{#2}
\providecommand{\BIBentrySTDinterwordspacing}{\spaceskip=0pt\relax}
\providecommand{\BIBentryALTinterwordstretchfactor}{4}
\providecommand{\BIBentryALTinterwordspacing}{\spaceskip=\fontdimen2\font plus
\BIBentryALTinterwordstretchfactor\fontdimen3\font minus
  \fontdimen4\font\relax}
\providecommand{\BIBforeignlanguage}[2]{{%
\expandafter\ifx\csname l@#1\endcsname\relax
\typeout{** WARNING: IEEEtran.bst: No hyphenation pattern has been}%
\typeout{** loaded for the language `#1'. Using the pattern for}%
\typeout{** the default language instead.}%
\else
\language=\csname l@#1\endcsname
\fi
#2}}
\providecommand{\BIBdecl}{\relax}
\BIBdecl

\bibitem{han2023high}
X.~Han, I.~R. Khan, and S.~Rahardja, ``{High Dynamic Range Image Tone Mapping:
  Literature review and performance benchmark},'' \emph{Digital Signal
  Processing}, p. 104015, 2023.

\bibitem{rana2018learning}
A.~Rana, G.~Valenzise, and F.~Dufaux, ``{Learning-based tone mapping operator
  for efficient image matching},'' \emph{IEEE Transactions on Multimedia},
  vol.~21, no.~1, pp. 256--268, 2018.

\bibitem{ak2022rv}
A.~Ak, A.~Goswami, W.~Hauser, P.~Le~Callet, and F.~Dufaux, ``{RV-TMO:
  Large-Scale Dataset for Subjective Quality Assessment of Tone Mapped
  Images},'' \emph{IEEE Transactions on Multimedia}, 2022.

\bibitem{wang2021deep}
L.~Wang and K.-J. Yoon, ``{Deep Learning for HDR Imaging: State-of-the-Art and
  Future Trends},'' \emph{IEEE transactions on pattern analysis and machine
  intelligence}, vol.~44, no.~12, pp. 8874--8895, 2021.

\bibitem{le2023single}
P.-H. Le, Q.~Le, R.~Nguyen, and B.-S. Hua, ``{Single-Image HDR Reconstruction
  by Multi-Exposure Generation},'' in \emph{Proceedings of the IEEE/CVF Winter
  Conference on Applications of Computer Vision}, 2023, pp. 4063--4072.

\bibitem{barua2023arthdr}
H.~B. Barua, G.~Krishnasamy, K.~Wong, K.~Stefanov, and A.~Dhall, ``{ArtHDR-Net:
  Perceptually Realistic and Accurate HDR Content Creation},'' in \emph{2023
  Asia Pacific Signal and Information Processing Association Annual Summit and
  Conference (APSIPA ASC)}.\hskip 1em plus 0.5em minus 0.4em\relax IEEE, 2023,
  pp. 806--812.

\bibitem{liu2020single}
Y.-L. Liu, W.-S. Lai, Y.-S. Chen, Y.-L. Kao, M.-H. Yang, Y.-Y. Chuang, and
  J.-B. Huang, ``{Single-Image HDR Reconstruction by Learning to Reverse the
  Camera Pipeline},'' in \emph{Proceedings of the IEEE/CVF Conference on
  Computer Vision and Pattern Recognition}, 2020, pp. 1651--1660.

\bibitem{sun2022putting}
Y.~Sun, W.~Liu, Q.~Bao, Y.~Fu, T.~Mei, and M.~J. Black, ``{Putting People in
  their Place: Monocular Regression of 3D People in Depth},'' in
  \emph{Proceedings of the IEEE/CVF Conference on Computer Vision and Pattern
  Recognition}, 2022, pp. 13\,243--13\,252.

\bibitem{lin2020cross}
K.~Lin, L.~Wang, K.~Luo, Y.~Chen, Z.~Liu, and M.-T. Sun, ``{Cross-domain
  complementary learning using pose for multi-person part segmentation},''
  \emph{IEEE Transactions on Circuits and Systems for Video Technology},
  vol.~31, no.~3, pp. 1066--1078, 2020.

\bibitem{kirillov2023segment}
A.~Kirillov, E.~Mintun, N.~Ravi, H.~Mao, C.~Rolland, L.~Gustafson, T.~Xiao,
  S.~Whitehead, A.~C. Berg, W.-Y. Lo \emph{et~al.}, ``{Segment anything},''
  \emph{arXiv preprint arXiv:2304.02643}, 2023.

\end{thebibliography}

%Biography text here without a photo.

%\begin{IEEEbiography}[{\includegraphics[width=1in,height=1.25in,clip,keepaspectratio]{}}]{IEEE Publications Technology Team}
%In this paragraph you can place your educational, professional background and research and other interests.
%\end{IEEEbiography}

\end{document}

\title{GTA-HDR: A Large-Scale Synthetic Dataset\\ for HDR Image Reconstruction\\{\Large Supplementary Material}}

\markboth{IEEE Transactions on Multimedia}{Barua \etal - GTA-HDR: A Large-Scale Synthetic Dataset for HDR Image Reconstruction}

\maketitle

%\IEEEPARstart{T}{his} additional supplementary document is an online appendix to the main article.

%\section{Research gap}
%\label{sec:gap}
%
%\begin{figure}[t]
%\centering
%\includegraphics[width=\linewidth]{images/research-gap.pdf}
%\caption{\textbf{.} .}
%\label{fig:tone_mapping}
%\end{figure}

%\section{Training strategies}
%\label{sec:res}
%
%\begin{figure}[t]
%\centering
%\includegraphics[width=\linewidth]%{images/finetuning.pdf}
%\caption{\textbf{.} .}
%\label{fig:tone_mapping}
%\end{figure}

\section{Inverse Tone Mapping}
\begin{figure}[t]
\centering
\includegraphics[width=\linewidth]{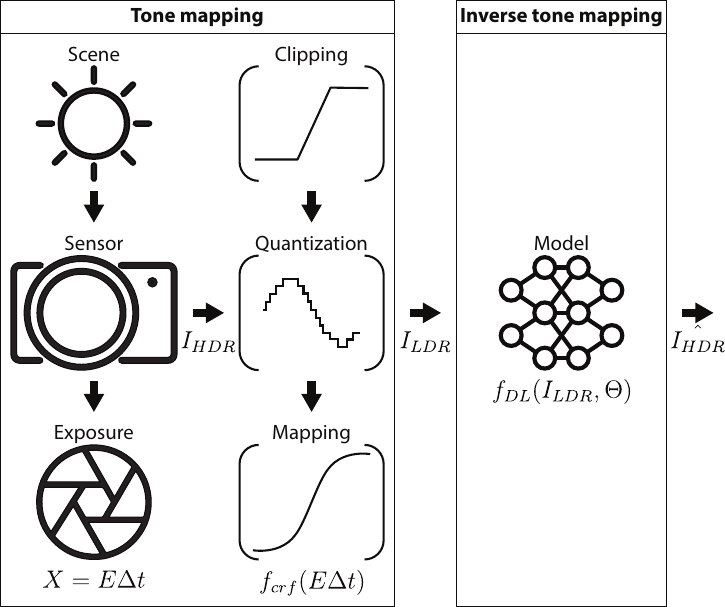}
\caption{\textbf{Tone mapping and inverse tone mapping processes.} The camera sensor function $X$ is the product of the sensor irradiance $E$ and exposure time $\Delta t$. The standard image formation pipeline (tone mapping) can be modeled with the function $f_{crf}(X)$, where $X = E\Delta t$. The goal of a data-driven inverse tone mapping model is to learn the function $f_{DL}(I_{LDR},\Theta)$, where $\Theta$ are the model parameters, which correctly approximates the inverse of $f_{crf}(X)$.}
\label{fig:tone_mapping}
\end{figure}

Tone mapping~\cite{han2023high,rana2018learning,ak2022rv}\mycomment{~\cite{han2023high,rana2019deep,rana2018learning,ak2022rv,alotaibi2023quality,han2023highnew,rana2018learning}} is the process of mapping the colors of HDR images capturing real-world scenes with a wide range of illumination levels to LDR images appropriate for standard displays with limited dynamic range.
Inverse tone mapping~\cite{wang2021deep} is the reverse process accomplished with either traditional (non-learning) methods or data-driven learning-based approaches.
Fig.~\ref{fig:tone_mapping} illustrates an overview of the tone mapping pipeline and the process of inverse tone mapping using a data-driven learning-based model. Here, $E$ is the sensor irradiance and $\Delta t$ is the exposure time.
The function $f_{crf}(E\Delta t)$ represents the tone mapping process, which outputs $I_{LDR}$ images given $I_{HDR}$ images captured by the camera sensor.
The main goal of any HDR image reconstruction technique is to reverse the tone mapping process using another function $f_{crf}^{-1}(I_{LDR})/\Delta t$, which outputs reconstructed $\hat{I_{HDR}}$ images given $I_{LDR}$ images.
The main challenge is that the steps in $f_{crf}(E\Delta t)$ are generally not reversible~\cite{le2023single}.
%Hence, to invert the non-linearities of this function, we need learning-based methods.
We can, however, approximate the reverse process with a data-driven learning-based model $f_{DL}(I_{LDR}, \Theta)$, which reconstructs $\hat{I_{HDR}}$ images given $I_{LDR}$ images, where $\Theta$ denotes the model parameters.
%Good generalization capabilities of the learning-based model require a significant amount of representative and diverse data for optimizing the model parameters $\theta$.

\section{HDR Image Reconstruction}
\begin{figure*}[t]
\centering
\includegraphics[width=0.95\linewidth]{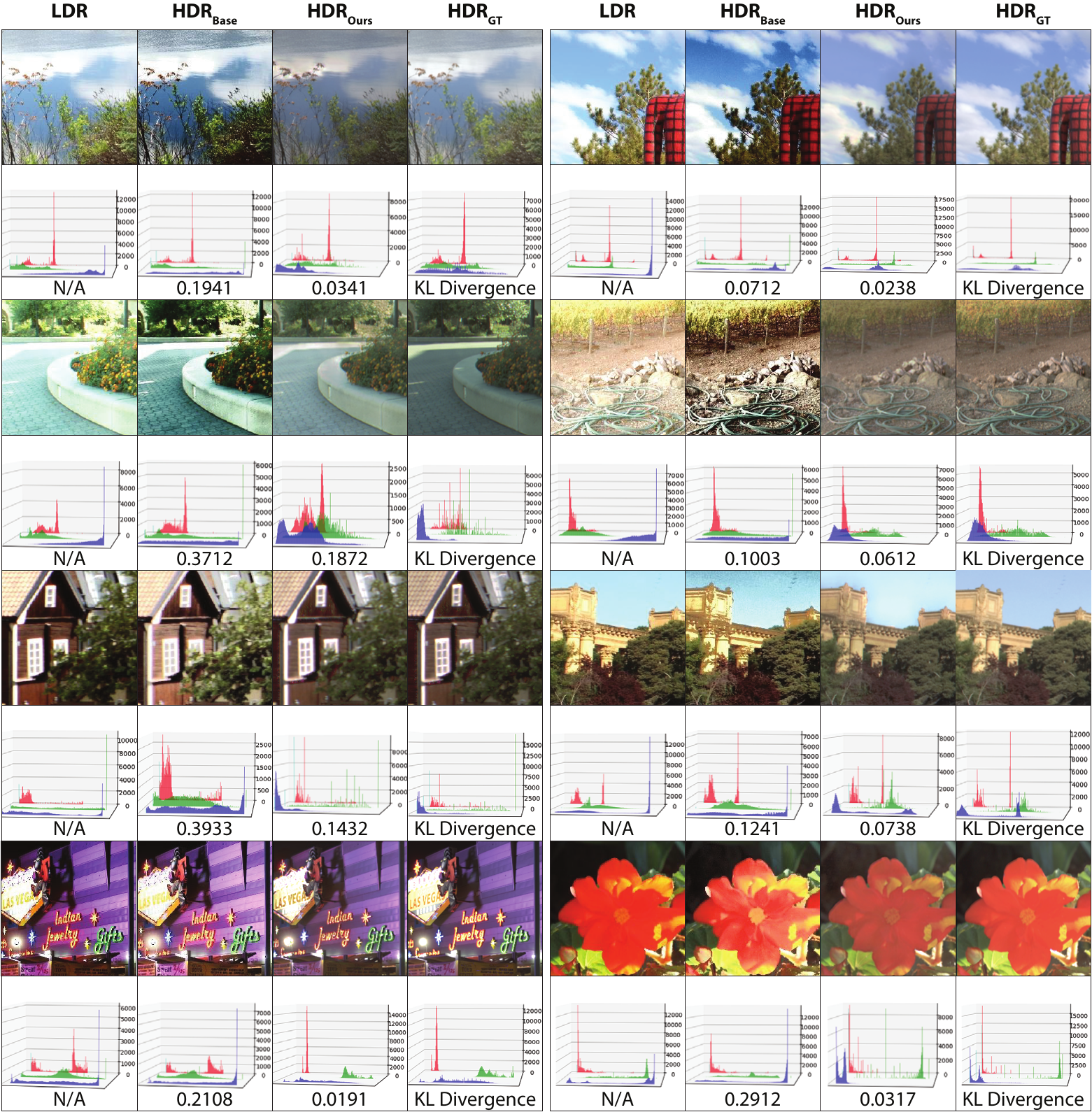}
\caption{\textbf{HDR images reconstructed with and without GTA-HDR as part of the training dataset, along with the RGB histograms and KL-divergence values.} \emph{Base}: HDR images reconstructed with ArtHDR-Net~\cite{barua2023arthdr} trained without GTA-HDR data; \emph{Ours}: HDR images reconstructed with ArtHDR-Net trained with GTA-HDR data; \emph{GT}: Ground truth.}
\label{fig:results_all_supp}
\end{figure*}

Fig.~\ref{fig:results_all_supp} illustrates examples of HDR images reconstructed by training  ArtHDR-Net~\cite{barua2023arthdr} with the GTA-HDR data in an end-to-end fashion.
The histograms of the ground truth and the reconstructed images are also included.
We can see that the histograms from the method trained with GTA-HDR data (\ie, HDR\textsubscript{Ours}) are more similar to the histograms of ground truth HDR images (\ie, HDR\textsubscript{GT}) than those from the method trained without GTA-HDR data (\ie, HDR\textsubscript{Base}).
We also report the Kullback-Leibler (KL)\mycomment{\footnote{https://en.wikipedia.org/wiki/Kullback\%E2\%80\%93Leibler_divergence}} divergence values for tone-mapped HDR\textsubscript{GT} and tone-mapped HDR\textsubscript{Base} and HDR\textsubscript{Ours} using the RGB intensities.
We see the average KL divergence of the RGB histogram intensity distributions are significantly lower for HDR\textsubscript{Ours} compared to HDR\textsubscript{Base}.

\section{Scene Diversity}
\begin{figure}[t]
\centering
\includegraphics[width=\linewidth]{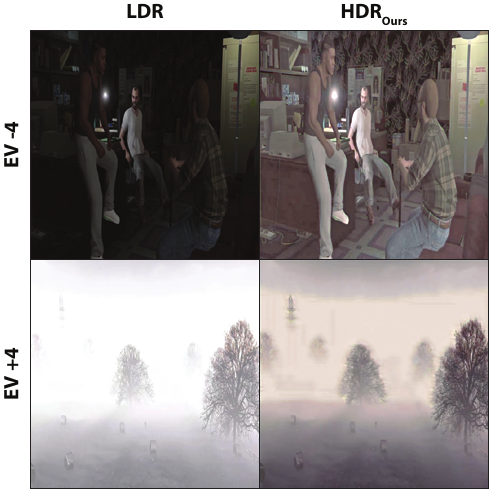}
\caption{\textbf{Performance of ArtHDR-Net~\cite{barua2023arthdr}.} The state-of-the-art method was trained with the GTA-HDR dataset and used for HDR image reconstruction from highly overexposed and underexposed synthetic LDR images. \emph{EV}: Exposure value; \emph{Ours}: HDR images reconstructed with ArtHDR-Net trained with GTA-HDR data.}
\label{fig:bias_qual1}
\end{figure}

\begin{figure}[t]
\centering
\includegraphics[width=\linewidth]{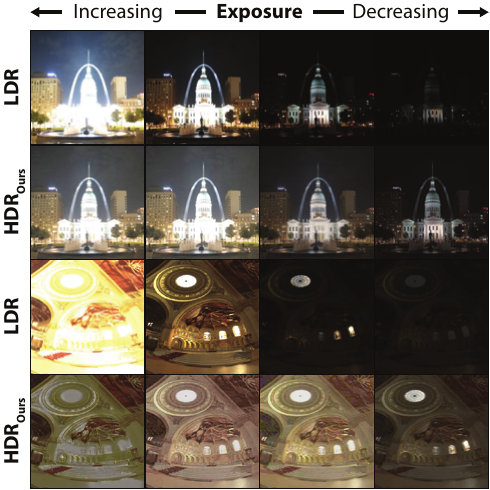}
\caption{\textbf{Performance of ArtHDR-Net~\cite{barua2023arthdr}.} The state-of-the-art method was trained with the GTA-HDR dataset and used for HDR image reconstruction from highly overexposed and underexposed real LDR images from the Internet. \emph{Ours}: HDR images reconstructed with ArtHDR-Net trained with GTA-HDR data.}
\label{fig:bias_qual2}
\end{figure}

\begin{figure}[t]
\centering
\includegraphics[width=\linewidth]{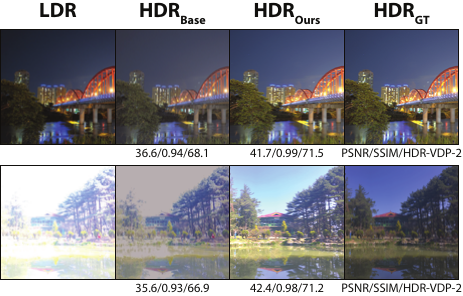}
\caption{\textbf{Performance of ArtHDR-Net~\cite{barua2023arthdr}.} We show the results on two extreme real in-the-wild images selected from HDR-Real~\cite{liu2020single} dataset. These images have extreme lighting conditions, color variations, and contrast levels. \emph{Base}: HDR images reconstructed with ArtHDR-Net trained without GTA-HDR data; \emph{Ours}: HDR images reconstructed with ArtHDR-Net trained with GTA-HDR data; \emph{GT}: Ground truth.}
\label{fig:in-the-wild}
\end{figure}

Fig.~\ref{fig:bias_qual1} illustrates further qualitative results from the state-of-the-art method ArtHDR-Net~\cite{barua2023arthdr} trained on GTA-HDR on extremely underexposed and overexposed images from a synthetic dataset.
Similarly, Fig.~\ref{fig:bias_qual2} demonstrates the performance for arbitrary real images from the Internet.
Both these cases show that GTA-HDR trained model is capable of recovering extremely over- and underexposed images with great fidelity.
To further illustrate the contribution of the GTA-HDR dataset on in-the-wild HDR image reconstruction, in Fig.~\ref{fig:in-the-wild} we show the results on two images selected from HDR-Real~\cite{liu2020single} dataset having extreme lighting, color, and contrast variations.
We also report the PSNR, SSIM, and HDR-VDP-2 scores. 
%We see that the HDR\textsubscript{Ours} outperforms  HDR\textsubscript{Base} both qualitatively and quantitatively.

\section{3D Human Pose and Shape Estimation}
\begin{figure*}[t]
\centering
\includegraphics[width=0.75\linewidth]{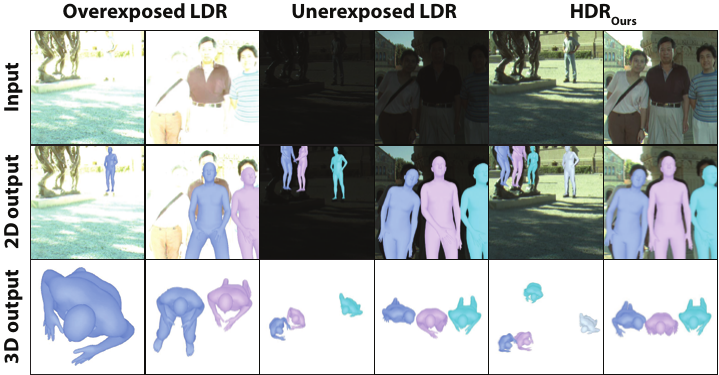}
\caption{\textbf{Impact of the GTA-HDR dataset on the performance of the state-of-the-art 3D human pose and shape estimation.} We used ArtHDR-Net~\cite{barua2023arthdr} trained with the GTA-HDR dataset for HDR image reconstruction from under- and overexposed LDR images. The resulting HDR images were used by BEV~\cite{sun2022putting} for 3D human pose and shape estimation. \emph{Ours}: HDR images reconstructed with ArtHDR-Net trained with GTA-HDR data.}
\label{fig:appl1}
\end{figure*}

%We compare the performances with original over-/under-exposed LDR images versus our generated HDR images.
We used BEV~\cite{sun2022putting} as a state-of-the-art pre-trained 3D human pose and shape estimator from images.
We tested the BEV model on the reconstructed HDR images from several versions of the state-of-the-art method ArtHDR-Net~\cite{barua2023arthdr}.
%and a variant trained with our GTA-HDR dataset (with E2E and FT strategies).
Fig.~\ref{fig:appl1} illustrates the contribution of GTA-HDR trained ArtHDR-Net using an end-to-end strategy for the task of 3D human pose and shape reconstruction.
The figure demonstrates that the performance of the BEV 3D pose estimator is better in the case of an ArtHDR-Net reconstructed HDR image (capturing $4$ people in the scene accurately) compared to the original LDR images (both under- and overexposed).

\section{Human Body Part Segmentation}
\begin{figure*}[t]
\centering
\includegraphics[width=0.75\linewidth]{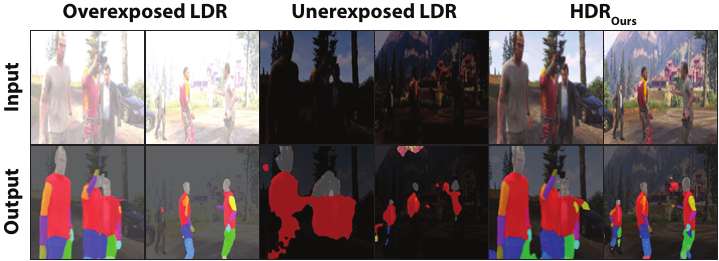}
\caption{\textbf{Impact of the GTA-HDR dataset on the performance of the state-of-the-art in 2D human body part segmentation.} We used ArtHDR-Net~\cite{barua2023arthdr} trained with the GTA-HDR dataset for HDR image reconstruction from under- and overexposed LDR images. The resulting HDR image was used by CDCL~\cite{lin2020cross} for 2D human body part segmentation. \emph{Ours}: HDR images reconstructed with ArtHDR-Net trained with GTA-HDR data.}
\label{fig:appl2}
\end{figure*}

In this experiment, we used CDCL~\cite{lin2020cross}, a state-of-the-art body part segmentation model.
Similar to the previous case, we tested the model on
reconstructed HDR images from several versions of ArtHDR-Net~\cite{barua2023arthdr}.
Fig.~\ref{fig:appl2} %provides an example of 
illustrates the contribution of the GTA-HDR dataset to this task.
The human body part segmentation results are more accurate with the reconstructed HDR images than the over- or underexposed LDR images.
For the overexposed LDR images, one person is completely missed in the second image.
For the underexposed LDR images, the output is noisy and erroneous. Finally, the processed HDRs using ArtHDR-Net~\cite{barua2023arthdr} trained with the GTA-HDR dataset deliver the most accurate segments.

\section{Semantic Segmentation}
\begin{figure*}[t]
\centering
\includegraphics[width=0.75\linewidth]{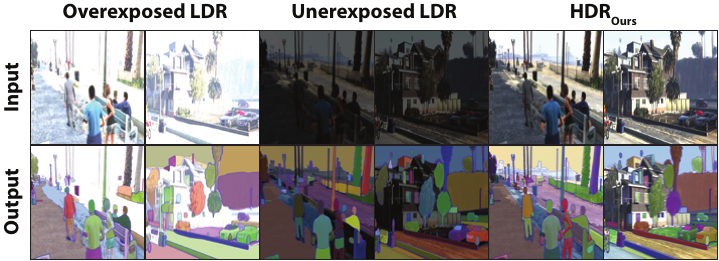}
\caption{\textbf{Impact of the GTA-HDR dataset on the performance of the state-of-the-art in semantic segmentation.} We used ArtHDR-Net~\cite{barua2023arthdr} trained with the GTA-HDR dataset for HDR image reconstruction from under- and overexposed LDR images. The resulting HDR image was used by SAM~\cite{kirillov2023segment} for semantic segmentation. \emph{Ours}: HDR images reconstructed with ArtHDR-Net trained with GTA-HDR data}
\label{fig:appl3}
\end{figure*}

Finally, we report an experiment on another vision task, \ie., holistic semantic segmentation of scenes, which is an important task in robotics and human-robot interaction.
We consider a recent state-of-the-art method called SAM~\cite{kirillov2023segment} as a pre-trained holistic scene segmentation model.
%We compare the output of SAM on HDRs generated by our training dataset (GTA-HDR with E2E) and strategy against normal unprocessed LDRs.
%We see that the accuracy of segmentation of complex objects in our HDRs is higher than the unprocessed LDRs.
Fig.~\ref{fig:appl3} illustrates the impact of the GTA-HDR trained model ArtHDR-Net~\cite{barua2023arthdr}.
The objects/buildings in the background are not segmented well in the overexposed LDR images.
Similarly, in the underexposed LDR images, even the near objects are sometimes segmented erroneously.
However, in the HDR images, these issues have been mitigated to a great extent.

\bibliographystyle{IEEEtran}
\bibliography{bibliography}